\documentclass[sigconf]{acmart}
\usepackage[utf8]{inputenc}
\usepackage[T1]{fontenc}
\usepackage{amsmath,amsfonts}
\usepackage{xcolor}
\usepackage{booktabs}
\usepackage{graphicx}
\usepackage{multirow}
\usepackage{array}
\usepackage{makecell}
\usepackage{enumitem}
\usepackage{adjustbox}
\usepackage{tabularx}
\usepackage{tikz}
\usetikzlibrary{shapes.geometric,arrows.meta,positioning,calc,shadows,backgrounds,fit}

\copyrightyear{2026}
\acmYear{2026}
\setcopyright{cc}
\setcctype{by}
\acmConference[GeoAI '26]{The 9th International Workshop on AI for Geographic Knowledge Discovery}{November 03--06, 2026}{Riverside, CA, USA}
\acmBooktitle{The 9th International Workshop on AI for Geographic Knowledge Discovery (GeoAI '26), November 03--06, 2026, Riverside, CA, USA}
\acmDOI{10.1145/3849726.3856091}
\acmISBN{979-8-4007-3052-8/2026/11}

\begin{document}

\title{GeoLEON-CLIP: Graph-Anchored Multimodal Contrastive Learning for Geographic Region Representation}


\author{Rados\l{}aw Ma\l{}ek}
\authornote{Both authors contributed equally to this research.}
\orcid{0009-0007-1834-9453}
\affiliation{%
  \institution{Wroc\l{}aw University of Science and Technology}
  \department{Department of Artificial Intelligence}
  \city{Wroc\l{}aw}
  \country{Poland}}
\email{radoslaw.malek@pwr.edu.pl}

\author{Szymon So\l{}tysiak}
\authornotemark[1]
\orcid{0009-0001-2054-2857}
\affiliation{%
  \institution{Wroc\l{}aw University of Science and Technology}
  \department{Department of Artificial Intelligence}
  \city{Wroc\l{}aw}
  \country{Poland}}
\email{szymon.soltysiak@pwr.edu.pl}

\author{Aleksandra Kawala-Sterniuk}
\orcid{0000-0001-7826-1292}
\affiliation{%
  \institution{Wroc\l{}aw University of Science and Technology}
  \department{Department of Artificial Intelligence}
  \city{Wroc\l{}aw}
  \country{Poland}}
\email{aleksandra.kawala-sterniuk@pwr.edu.pl}

\author{Piotr Szymański}
\orcid{0000-0002-7733-3239}
\affiliation{%
  \institution{Wrocław University of Science and Technology}
  \department{Department of Artificial Intelligence}
  \city{Wrocław}
  \country{Poland}
}
\email{piotr.szymanski@pwr.edu.pl}

\begin{abstract}
Geographic regions are naturally multimodal: satellite imagery captures visual morphology, textual descriptions encode human-interpretable semantics, and map-derived structures describe functional urban context. Existing geospatial contrastive learning methods often align image–text or image–location pairs, leaving structured map information underused. We introduce GeoLEON-CLIP, a graph-anchored multimodal contrastive framework for learning reusable geographic region representations. The model aligns three views of the same H3-indexed region: an OpenStreetMap-derived graph embedding, a satellite image crop, and a textual region description. The graph branch uses frozen pretrained LEON embeddings and, as we show empirically, acts as a semantic anchor between visual appearance and language. Frozen graph, image, and text encoders are connected through trainable residual projection heads and optimized with a symmetric tri-modal contrastive objective. Experiments show strong graph–image and graph–text alignment, with graph–image retrieval achieving the best cross-modal performance in the mixed-resolution pool. To test the anchoring hypothesis directly, we introduce chained cross-modal recall $\mathrm{R}@A@B$, which routes retrieval through the graph modality and compares the result with direct retrieval at a matched candidate budget. In the strictest setting, $\mathrm{R}@1@1$, routing image$\rightarrow$graph$\rightarrow$text and text$\rightarrow$graph$\rightarrow$image is more accurate than direct image--text retrieval by 4.8 and 4.4 points respectively, even though the chain must commit to a single intermediate cell. Downstream probes show task-dependent gains, with the strongest improvements on Airbnb and Beijing housing. The results suggest that OSM-derived graph structure can serve as an effective anchor for multimodal geospatial representation learning that combines maps, imagery, and language.
\end{abstract}

\ccsdesc[500]{Computing methodologies~Neural networks}
\ccsdesc[500]{Applied computing~Geographic information systems}
\ccsdesc[300]{Information systems~Spatial-temporal systems}

\keywords{geospatial representation learning, self-supervised learning, graph masked autoencoders, OpenStreetMap, H3, multimodal contrastive learning, GeoAI}

\maketitle

\section{Introduction}
\par Modern geospatial analysis increasingly depends on reusable representations of locations rather than one-off, task-specific features \cite{goodchild2004gis,boutayeb2024comprehensive,chen2024self}. Geographic regions are inherently multimodal: satellite imagery captures physical morphology and land-cover patterns, volunteered geographic information describes functional urban structure, mobility traces reflect human activity, and textual descriptions encode human-interpretable place semantics \cite{jean2016poverty,openstreetmap2008,mai2024opportunities}.
\par Recent geospatial foundation models and contrastive pretraining approaches have shown that such heterogeneous data can be used to learn transferable spatial representations \cite{klemmer2025satclip,vivancocepeda2023geoclip,yan2024urbanclip}. However, adapting these representations to downstream geographic prediction tasks still often depends on scarce, spatially uneven, and task-specific labels, which remain a practical bottleneck for robust generalization across cities, administrative definitions, and spatial scales \cite{boutayeb2024comprehensive,hao2026geospatial}.
\par Self-supervised and contrastive learning offer a way to reduce this dependence on labeled data by exploiting naturally co-located geographic observations \cite{simclr2020,radford2021learning,wang2020understanding,chen2024self}. 
\par Yet structured map-derived information, such as OpenStreetMap-based urban function and neighborhood topology, is often treated as an auxiliary feature rather than as a central modality \cite{openstreetmap2008,haklay2010vgi,wozniak2021hex2vec,regiondcl2023}. This leaves an open question: can graph representations derived from volunteered geographic information provide a stable semantic anchor for aligning visual and textual views of the same place?
\par This paper builds on LEON, a self-supervised graph masked autoencoder for urban region representation learning \cite{soltysiak2026leon}. LEON represents cities as H3-indexed graphs, connects neighboring H3 cells through spatial adjacency, and derives node features from OpenStreetMap amenity counts \cite{uberH3,sahr2003dggs,openstreetmap2008,hou2022graphmae,velickovic2018graph}. By reconstructing masked node attributes, the model learns embeddings that encode both local urban function and surrounding spatial context. Motivated by the strong graph-only representations obtained in this earlier work, we investigate whether the LEON embedding space can be extended from a unimodal graph representation into a shared multimodal geo-semantic space.
\par We propose GeoLEON-CLIP, a graph-anchored multimodal contrastive framework that aligns three views of the same H3 cell: an OpenStreetMap-derived graph embedding, a satellite image crop, and a textual region description. The architecture is intentionally modular. The graph branch uses the pretrained LEON encoder, the image branch uses a frozen visual foundation encoder, and the text branch uses a frozen multilingual embedding model. During alignment, only lightweight residual projection heads and the contrastive logit scale are trained. This design isolates the contribution of multimodal alignment while preserving the reusable structure learned by the graph encoder.
\par
\begin{figure*}[t]
    \centering
    \begingroup
    \sffamily
    \scriptsize

    \newcommand{\GeoLEONFigImage}[3]{%
        \IfFileExists{#3}{%
            \includegraphics[width=#1,height=#2]{#3}%
        }{%
            \fbox{%
                \begin{minipage}[c][#2][c]{#1}
                    \centering\scriptsize\texttt{\detokenize{#3}}
                \end{minipage}%
            }%
        }%
    }

    \newcommand{\GeoLEONGridCube}[3]{%
        \def\cubesize{0.70}%
        \def\cubespacing{0.82}%
        \pgfmathsetmacro{\px}{(#1-1) * \cubespacing * (-0.866) + (#2-1) * \cubespacing * (0.866)}%
        \pgfmathsetmacro{\py}{(#1-1) * \cubespacing * (-0.5) + (#2-1) * \cubespacing * (-0.5) + (#3-1) * \cubespacing * (1)}%
        \begin{scope}[shift={(\px,\py)}]
            \pgfmathsetmacro{\dx}{\cubesize * 0.866}%
            \pgfmathsetmacro{\dy}{\cubesize * 0.5}%
            \draw[fill=red!40, thick, draw=black!80, line join=round]
                (0,\cubesize) -- (-\dx,\cubesize+\dy) -- (0,\cubesize+2*\dy) -- (\dx,\cubesize+\dy) -- cycle;
            \node[text=black!90] at (0,\cubesize+\dy) {\scalebox{0.95}{$\mathbf{G}'$}};
            \draw[fill=green!40, thick, draw=black!80, line join=round]
                (0,0) -- (-\dx,\dy) -- (-\dx,\cubesize+\dy) -- (0,\cubesize) -- cycle;
            \node[text=black!90] at (-0.5*\dx,0.5*\cubesize+0.5*\dy) {\scalebox{0.95}{$\mathbf{T}'$}};
            \draw[fill=yellow!40, thick, draw=black!80, line join=round]
                (0,0) -- (\dx,\dy) -- (\dx,\cubesize+\dy) -- (0,\cubesize) -- cycle;
            \node[text=black!90] at (0.5*\dx,0.5*\cubesize+0.5*\dy) {\scalebox{0.95}{$\mathbf{I}'$}};
        \end{scope}%
    }

    \resizebox{0.98\textwidth}{!}{%
    \begin{tikzpicture}[
        data_box/.style={
            rectangle,
            rounded corners=3pt,
            draw=black!60,
            thick,
            fill=white,
            minimum width=2.8cm,
            minimum height=1.65cm,
            align=center,
            inner sep=0pt,
            drop shadow={opacity=0.18, shadow xshift=1pt, shadow yshift=-1pt}
        },
        model_box/.style={
            rectangle,
            rounded corners=5pt,
            draw=black!80,
            thick,
            fill=gray!10,
            minimum width=2.8cm,
            minimum height=1.65cm,
            align=center,
            inner sep=3pt,
            drop shadow={opacity=0.18, shadow xshift=1pt, shadow yshift=-1pt}
        },
        embedding_vec/.style={
            rectangle,
            draw=black!80,
            thick,
            minimum width=0.58cm,
            minimum height=1.65cm,
            align=center,
            font=\small\bfseries,
            drop shadow={opacity=0.18, shadow xshift=1pt, shadow yshift=-1pt}
        },
        arrow_style/.style={->, >=Stealth, thick, color=black!80},
        paired_arrow/.style={<->, >=Stealth, thick, dashed, color=black!75},
        label_text/.style={font=\footnotesize\sffamily, align=center, color=black!85},
        title_text/.style={font=\bfseries\sffamily, align=center, color=black}
    ]

    \node[data_box, fill=gray!10, xshift=-0.16cm, yshift=0.16cm] at (0,3.45) {\phantom{\rule{2.68cm}{1.53cm}}};
    \node[data_box, fill=gray!5, xshift=-0.08cm, yshift=0.08cm] at (0,3.45) {\phantom{\rule{2.68cm}{1.53cm}}};
    \node[data_box, fill=white] (web_docs) at (0,3.45) {\GeoLEONFigImage{2.68cm}{1.53cm}{example_page.png}};
    \node[label_text, above=0.16cm of web_docs] {\textbf{Web sources}\\Wikipedia, Wikivoyage, Reddit};

    \node[data_box] (text_data) at (3.75,3.45) {\parbox{2.30cm}{\centering\tiny A transit-oriented district near Yatsuka Station with everyday mobility functions.}};
    \node[label_text, above=0.16cm of text_data] {\textbf{Cell-level text}\\H3 aggregation};

    \node[model_box] (jina_model) at (7.55,3.45) {\textbf{Jina embeddings v3}\\\footnotesize chunk mean + 128D MRL};
    \node[embedding_vec, fill=green!40] (emb_text) at (10.75,3.45) {$\mathbf{T}$};

    \draw[arrow_style] (web_docs) -- (text_data);
    \draw[arrow_style] (text_data) -- (jina_model);
    \draw[arrow_style] (jina_model) -- (emb_text);

    \node[data_box, fill=gray!10, xshift=-0.16cm, yshift=0.16cm] at (0,0) {\phantom{\rule{2.68cm}{1.53cm}}};
    \node[data_box, fill=blue!5, xshift=-0.08cm, yshift=0.08cm] at (0,0) {\phantom{\rule{2.68cm}{1.53cm}}};
    \node[data_box, fill=white] (input_map) at (0,0) {\GeoLEONFigImage{2.68cm}{1.53cm}{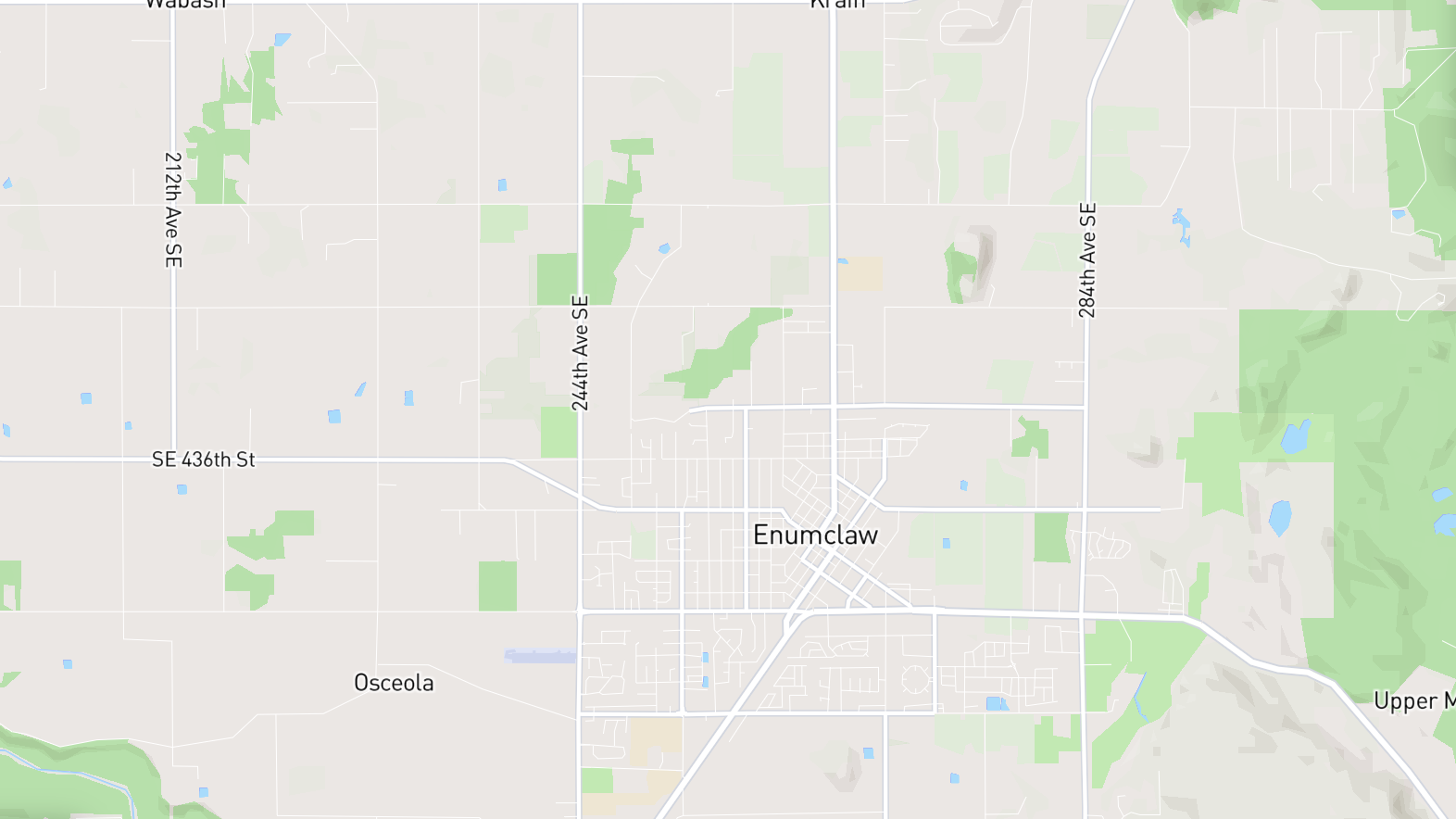}};
    \node[label_text, above=0.16cm of input_map] {\textbf{OSM layers}\\functional objects};

    \node[data_box, fill=yellow!10] (hex_grid) at (3.75,0) {\GeoLEONFigImage{2.68cm}{1.53cm}{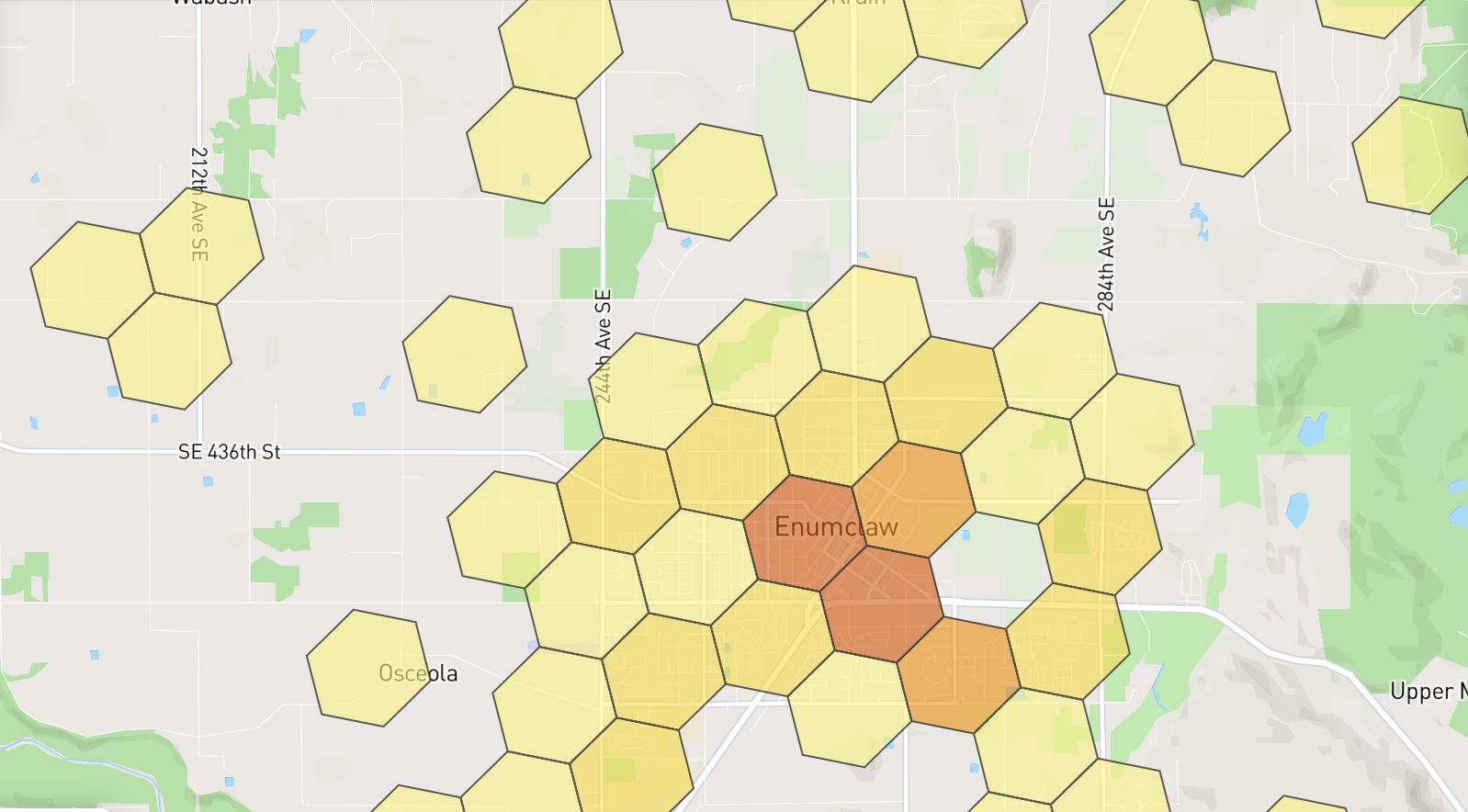}};
    \node[label_text, above=0.16cm of hex_grid] {\textbf{H3 graph}\\spatial adjacency};

    \node[model_box] (leon_model) at (7.55,0) {\textbf{LEON encoder}\\\footnotesize graph masked autoencoder};
    \node[embedding_vec, fill=red!40] (emb_graph) at (10.75,0) {$\mathbf{G}$};

    \draw[arrow_style] (input_map) -- (hex_grid);
    \draw[arrow_style] (hex_grid) -- (leon_model);
    \draw[arrow_style] (leon_model) -- (emb_graph);

    \node[data_box, fill=gray!10, xshift=-0.16cm, yshift=0.16cm] at (0,-3.45) {\phantom{\rule{2.68cm}{1.53cm}}};
    \node[data_box, fill=blue!5, xshift=-0.08cm, yshift=0.08cm] at (0,-3.45) {\phantom{\rule{2.68cm}{1.53cm}}};
    \node[data_box, fill=white] (sat_img) at (0,-3.45) {\GeoLEONFigImage{2.68cm}{1.53cm}{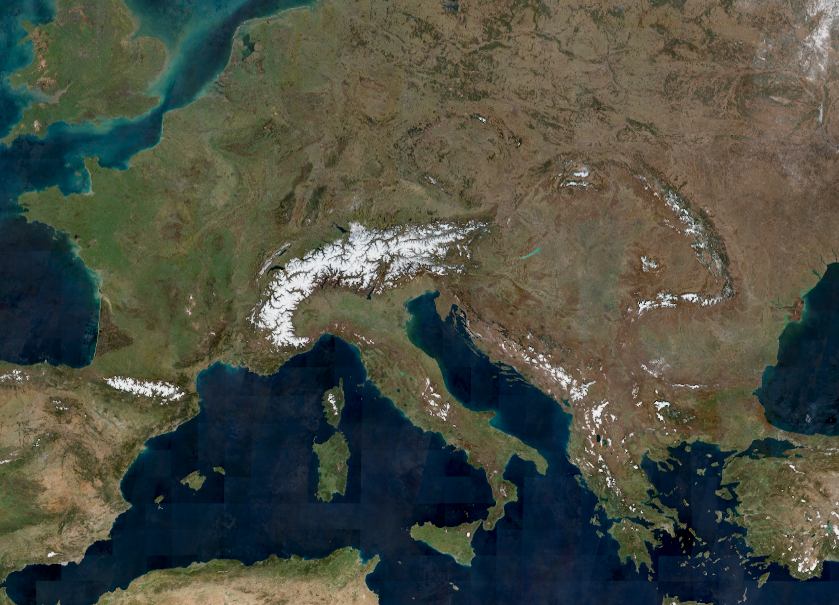}};
    \node[label_text, above=0.16cm of sat_img] {\textbf{Sentinel-2 RGB}\\satellite imagery};

    \node[data_box] (sat_crop) at (3.75,-3.45) {\GeoLEONFigImage{2.68cm}{1.53cm}{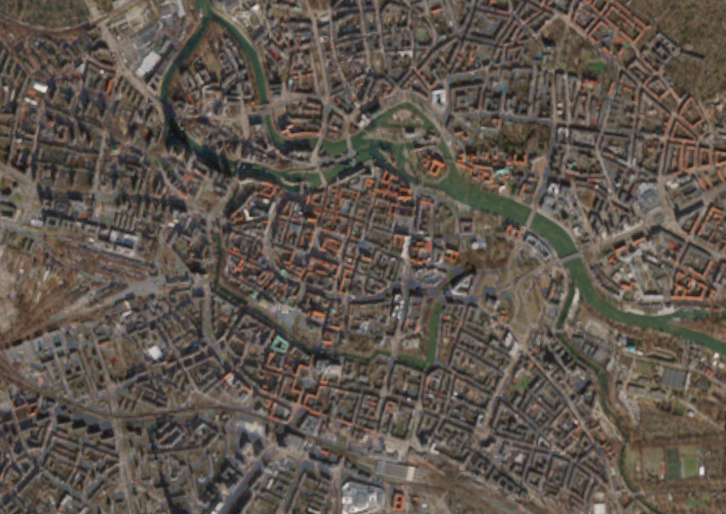}};
    \node[regular polygon, regular polygon sides=6, draw=orange, very thick, minimum size=1.05cm] at (sat_crop.center) {};
    \node[label_text, above=0.16cm of sat_crop] {\textbf{Image crop}\\central patch pooling};

    \node[model_box] (dino_model) at (7.55,-3.45) {\textbf{DINOv3 ViT-L/16}\\\footnotesize frozen satellite encoder};
    \node[embedding_vec, fill=yellow!40] (emb_image) at (10.75,-3.45) {$\mathbf{I}$};

    \draw[arrow_style] (sat_img) -- (sat_crop);
    \draw[arrow_style] (sat_crop) -- (dino_model);
    \draw[arrow_style] (dino_model) -- (emb_image);

    \draw[paired_arrow] (text_data.south) -- node[right=0.05cm, label_text] {} (hex_grid.north);
    \draw[paired_arrow] (hex_grid.south) -- node[right=0.05cm, label_text] {} (sat_crop.north);

    \node[draw=black!50, dashed, thick, rounded corners, fill=purple!5,
          minimum width=7.55cm, minimum height=8.45cm] (contrastive_space) at (15.25,0) {};
    \node[title_text, above=0.16cm of contrastive_space] {\Large \textbf{GeoLEON-CLIP}};
    \node[label_text, below=0.16cm of contrastive_space] {Shared geo-semantic space trained with symmetric InfoNCE};

    \node[model_box, fill=orange!20, minimum width=1.35cm, minimum height=3.75cm] (proj) at (12.75,0)
        {\rotatebox{90}{\textbf{Residual projection heads}}};

    \draw[arrow_style] (emb_text.east) to[out=0,in=180] (proj.west);
    \draw[arrow_style] (emb_graph.east) -- (proj.west);
    \draw[arrow_style] (emb_image.east) to[out=0,in=180] (proj.west);
    \draw[-{Stealth[scale=1.2]}, line width=2.2pt, color=black!80] (proj.east) -- ++(0.70,0);

    \begin{scope}[shift={(16.15,-0.30)}]
        \foreach \z in {0,1,2} {
            \foreach \x in {0,1,2} {
                \foreach \y in {0,1,2} {
                    \GeoLEONGridCube{\x}{\y}{\z}
                }
            }
        }
    \end{scope}

    \end{tikzpicture}%
    }
    \endgroup
    \caption{Overview of the GeoLEON-CLIP architecture. Each H3 anchor is represented by three paired views: an OSM-derived graph view encoded by the pretrained LEON graph masked autoencoder, a Sentinel-2 image view encoded by a frozen DINOv3 ViT-L/16 satellite encoder, and a text view encoded by frozen Jina embeddings. Trainable residual projection heads map the three frozen encoder outputs into a shared geo-semantic space optimized with symmetric pairwise contrastive losses.}
    \label{fig:geoleon_architecture}
\end{figure*}

 Figure~\ref{fig:geoleon_architecture} provides an overview of the proposed architecture. Each H3 anchor defines a paired graph--image--text training example. The graph view is derived from OSM features aggregated over the H3 grid, the image view is extracted from Sentinel-2 crops using central DINOv3 patch pooling, and the text view is encoded from Wikipedia, Wikivoyage, and Reddit descriptions. The three frozen encoders are connected through trainable projection heads and optimized with a symmetric tri-modal contrastive objective.
\par GeoLEON-CLIP targets a gap between graph-based urban representation learning and CLIP-style geospatial pretraining. Whereas prior contrastive models often align imagery with text, coordinates, or location encodings, our approach explicitly asks whether an OSM-derived graph representation can act as a bridge between visual morphology and human-readable place semantics. We evaluate this hypothesis through pairwise and chained cross-modal retrieval,
modality-geometry analysis, projection-head ablations, and downstream spatial
prediction probes.
The main contributions of this paper are as follows:
\begin{enumerate}[leftmargin=*]
\item \textbf{Graph-anchored multimodal geospatial alignment.} We introduce GeoLEON-CLIP, a modular architecture that aligns OSM-derived graph structure, satellite imagery, and textual place semantics into a shared H3-indexed embedding space.
\item \textbf{H3-paired multimodal dataset construction.} We describe a multi-scale dataset design over global metropolitan areas in which graph, image, and text views are paired through common H3 cell identifiers at district, neighborhood, and block scales.
\item \textbf{Contrastive extension of LEON.} We extend the self-supervised LEON graph encoder from graph-only representation learning to CLIP-style multimodal alignment, testing whether graph embeddings can serve as a semantic anchor between visual and textual modalities.
\item \textbf{Chained cross-modal recall.} We introduce $\mathrm{R}@A@B$, a budget-matched two-hop retrieval metric that measures how much of the image--text relation is recoverable by routing through the graph modality. It turns the informal notion of a ``semantic anchor'' into a falsifiable comparison against direct retrieval, isolates a regime, $A=B=1$, in which the detour through the graph is strictly better than the direct path, and includes alternative bridge placements as a control.
\item \textbf{Evaluation of graph-anchored multimodal alignment.} We evaluate GeoLEON-CLIP with pairwise and chained retrieval, geographic-error and shared-space diagnostics, projection-head and training ablations, and downstream Ridge probes across five geographic prediction tasks.
\end{enumerate}


\section{Related Work}
\label{sec:related_work}
\par Existing work has explored several forms of geospatial alignment, including image--location pretraining, image--text alignment for urban region profiling, and more recent multi-modal geospatial contrastive frameworks \cite{klemmer2025satclip,vivancocepeda2023geoclip,liu2024remoteclip,zhang2024rs5m,wang2024skyscript,yan2024urbanclip}.
\paragraph{Spatial and urban representation learning.}
Spatial representation learning aims to encode places, regions, or grid cells into dense vectors that preserve functional, morphological, and contextual properties of geographic space \cite{goodchild2004gis,boutayeb2024comprehensive,hao2026geospatial,chen2024self}. Early urban prediction pipelines typically relied on hand-crafted GIS indicators such as POI densities, land-use proportions, road-network statistics, accessibility measures, and neighborhood aggregates \cite{marti2021neighborhood,openstreetmap2008}. 
\par More recent approaches learn region representations directly from spatial data sources, including volunteered geographic information, satellite imagery, mobility traces, and spatial graphs. H3 provides a practical hierarchical indexing system for representing geographic space as multi-resolution hexagonal cells \cite{uberH3, sahr2003dggs}. Within this line of work, Hex2Vec learns context-aware embeddings of H3 cells from OpenStreetMap tags using a Skip-gram-style objective \cite{wozniak2021hex2vec}, while S2Vec uses self-supervised masked autoencoding over rasterized built-environment features in S2 cells \cite{s2vec2026}. These methods demonstrate that map-derived attributes can provide useful task-agnostic spatial representations, but they remain largely unimodal and do not explicitly align map structure with visual or textual views of the same region.
\paragraph{Masked autoencoding for graphs and spatial structure.}
Masked prediction has become a central self-supervised learning paradigm across language, vision, and graph learning. BERT introduced masked language modeling for bidirectional language representation learning \cite{devlin2019bert}, while MAE showed that reconstructing masked image patches can scale effectively for visual representation learning \cite{he2022masked}. In graph learning, GAT introduced attention-based neighborhood aggregation \cite{velickovic2018graph}, and GraphMAE adapted the masked autoencoding principle to graphs by masking node attributes and reconstructing them from graph context using a scaled cosine error objective \cite{hou2022graphmae}. LEON follows this line of work and instantiates graph masked autoencoding in a geographic setting: urban regions are represented as H3-indexed spatial graphs, edges encode cell adjacency, and node attributes are derived from OpenStreetMap features. This adaptation is important because geographic graphs differ from standard graph benchmarks: their edges correspond to physical proximity, node attributes are sparse and highly skewed, and nearby regions are strongly spatially autocorrelated. 
\paragraph{Vision-language and geospatial contrastive learning.}
Contrastive language-image pretraining has shown that large-scale image-text alignment can produce transferable visual representations \cite{radford2021learning}. This idea has recently been adapted to remote sensing and geospatial representation learning. SatCLIP learns global location embeddings by contrastively aligning satellite imagery with geographic coordinates \cite{klemmer2025satclip}, while GeoCLIP aligns images with continuous GPS representations for worldwide geolocalization \cite{vivancocepeda2023geoclip}. RemoteCLIP and GeoRSCLIP specialize vision-language alignment for remote sensing by training or adapting CLIP-style models on remote-sensing image--text pairs \cite{liu2024remoteclip,zhang2024rs5m}. SkyScript constructs a large remote-sensing vision-language dataset by connecting unlabeled remote-sensing imagery with OpenStreetMap-derived semantic descriptions \cite{wang2024skyscript}. UrbanCLIP is particularly relevant to urban representation learning because it uses generated textual descriptions to enhance satellite-image-based urban region profiling \cite{yan2024urbanclip}. These works demonstrate the value of contrastive alignment for geographic imagery and language, but most of them align either image--text, image--location, or image--caption pairs. Structured OSM-derived graph representations are usually absent or used indirectly as a source of textual semantics rather than as a dedicated modality.
\paragraph{Foundation encoders for modular multimodal alignment.}
Recent foundation encoders make it possible to build modular multimodal systems without training every component from scratch. In computer vision, DINOv3 provides strong self-supervised visual backbones, including variants trained on large-scale satellite imagery \cite{simeoni2025dinov3}. For text, multilingual embedding models such as Jina Embeddings V3 provide compact semantic representations across languages and tasks through task-specific low-rank adaptations \cite{sturua2025jina}. GeoLEON-CLIP follows this modular design: the graph branch uses frozen pretrained LEON embeddings, while the image and text encoders are also frozen. Training is restricted to lightweight residual projection heads and the contrastive logit scale. This design isolates the alignment problem from full encoder pretraining and allows us to test whether graph-derived geographic structure can organize visual and textual semantics in a shared embedding space.
\paragraph{Positioning of GeoLEON-CLIP}
Compared with spatial embedding methods such as Hex2Vec and S2Vec, our model is explicitly multimodal. Compared with remote-sensing vision-language models such as RemoteCLIP, GeoRSCLIP, SkyScript, and UrbanCLIP, our model introduces an explicit OSM-derived graph branch rather than relying only on images, text, or coordinates. Compared with location-image models such as SatCLIP and GeoCLIP, our method aligns region-level graph, image, and text views through a common H3 cell identifier. The central hypothesis is therefore not simply that multimodal pretraining helps geographic prediction, but that OSM-derived graph structure can act as a semantic anchor between overhead visual morphology and human-readable place descriptions. Table~\ref{tab:related_work_positioning} in the appendix summarizes these differences side by side.

\section{Tri-Modal H3 Dataset}
\label{sec:dataset}
%
\par The Tri-Modal H3 Dataset is a fully paired collection anchored to Uber's H3 hexagonal grid, aligning OpenStreetMap graph features, Sentinel-2 imagery, and textual descriptions at consistent geographic locations.
\subsection{Spatial Anchors and Study Areas}
\par The dataset is organized around H3 cells, which provide a hierarchical hexagonal indexing system for representing geographic space at multiple spatial resolutions \cite{uberH3}. Each example is anchored by an H3 index $h_i$ and contains three paired modalities:
\begin{equation}
    a_i = (g_i, v_i, t_i, h_i),
\end{equation}
where {$g_i$} denotes the graph-derived view, $v_i$ the satellite image view, and $t_i$ the textual view. The H3 index is the common alignment key used to join OpenStreetMap-derived graph features, Sentinel-2 image crops, textual descriptions, and downstream tabular observations.
\par Unlike weakly paired multimodal datasets in which one or more modalities may be missing, our dataset is fully paired at every spatial resolution used in this work, so the number of complete graph--image--text triplets equals the number of H3 anchors at each resolution.
\par The current dataset covers {$35$} metropolitan areas across six inhabited continents, selected to expose the model to varied urban morphology, including dense transit corridors, monocentric historical centers, polycentric megacities, industrial belts, coastal regions, and sparse urban peripheries. We construct the dataset at three H3 resolutions, corresponding approximately to district, neighborhood, and block scales.
\par Table \ref{tab:h3_resolutions} summarizes the H3 resolutions used in GeoLEON-CLIP. Area values follow the official H3 cell statistics, with lower resolution numbers corresponding to larger cells (district scale) and higher resolution numbers to smaller cells (block scale).
\begin{table}[t]
\centering
\caption{H3 resolutions used in GeoLEON-CLIP. Average areas follow the official H3 cell statistics.}
\label{tab:h3_resolutions}
\small
\setlength{\tabcolsep}{4pt}
\renewcommand{\arraystretch}{1.12}
\begin{tabular}{@{}l r r p{0.28\columnwidth}@{}}
\toprule
\textbf{Resolution} & \textbf{Avg. area} & \textbf{Anchors} & \textbf{Interpretation} \tabularnewline
\midrule
7 & 5.16 km\textsuperscript{2} & 17,104 & district / macro urban structure \tabularnewline
8 & 0.74 km\textsuperscript{2} & 119,791 & neighborhood-scale context \tabularnewline
9 & 0.10 km\textsuperscript{2} & 838,537 & block-scale POI and morphology \tabularnewline
\bottomrule
\end{tabular}
\end{table}
\par Table \ref{tab:dataset_components} summarizes the provenance, alignment key, and temporal scope of each modality.
\begin{table*}[t]
\centering
\caption{Dataset components, provenance, and temporal scope. All retained H3 anchors are fully paired across graph, image, and text modalities.}
\label{tab:dataset_components}
\scriptsize
\setlength{\tabcolsep}{4pt}
\renewcommand{\arraystretch}{1.15}
\begin{tabular}{@{}p{0.11\textwidth}p{0.19\textwidth}p{0.15\textwidth}p{0.15\textwidth}p{0.33\textwidth}@{}}
\toprule
\textbf{Modality} & \textbf{Source} & \textbf{Alignment key} & \textbf{Coverage} & \textbf{Temporal scope and statistics} \tabularnewline
\midrule
Graph & OpenStreetMap objects aggregated to H3 & H3 index & 100\% of anchors & OSM snapshot from December 2025; $K=31$ functional categories used as graph node features. \tabularnewline
Image & Sentinel-2 Level-2A RGB crops & H3 polygon / centroid & 100\% of anchors & Primary acquisition window: March 2025--February 2026; scenes closest to December 2025 preferred; approximately 5\% of crops use the nearest valid observation outside the primary window. \tabularnewline
Text & Textual descriptions associated with OSM-derived objects and external web sources & H3 cell or parent H3 cell & 100\% of anchors & OSM object set fixed to the December 2025 snapshot; external textual content accessed in March 2026; long documents are chunked and averaged before alignment. \tabularnewline
\bottomrule
\end{tabular}
\end{table*}
\subsection{Graph Modality}
\par The graph modality is derived from OpenStreetMap objects observed in the December 2025 snapshot. For each H3 cell, OSM objects are assigned to a single H3 anchor and aggregated into {$K=31$} functional categories. We use all OSM geometry types available for the selected categories. Point objects are assigned directly using their coordinates. For ways and relations, including polygonal geometries, we compute the geometric centroid and assign the object to the H3 cell containing that centroid. This centroid-based assignment ensures that each OSM object contributes to at most one H3 cell and prevents large polygons from dominating multiple neighboring cells. If a centroid lies exactly on the boundary between H3 cells, the object is assigned uniformly at random to one of the adjacent cells to avoid duplicate counting.
\par Let {$a_{ik}$} denote the raw count of OSM objects from category {$k$} assigned to cell {$h_i$}. Since OSM counts are sparse and highly skewed across dense and peripheral urban regions, raw counts are log-transformed \eqref{eq1}:
\begin{equation}
    c_{ik} = \log(1 + a_{ik}).
    \label{eq1}
\end{equation}
\par A binary empty-cell indicator $b_i \in \{0,1\}$ is appended to distinguish structurally added empty nodes from active cells with low observed density. The final node feature is as follows \eqref{eq2}:
\begin{equation}
    x_i = [c_i \parallel b_i] \in \mathbb{R}^{K+1}.
    \label{eq2}
\end{equation}
\par Edges are added symmetrically between immediate H3 neighbors \eqref{eq3}:
\begin{equation}
    (v_i, v_j) \in \mathcal{E}
    \Longleftrightarrow
    i \neq j \land \mathrm{dist}_{H3}(h_i,h_j)=1.
    \label{eq3}
\end{equation}
\par This construction produces a spatial adjacency graph in which message passing follows physical proximity between neighboring cells. For a complete audit of graph construction, we report the number of retained cells, number of edges, mean node degree, and non-empty node rate for each H3 resolution in the final dataset release.
\par The {$K=31$} tags behind these categories span health, retail, food and drink, tourism, land use, leisure, buildings, transport, and culture and education; they are listed in full in Table~\ref{tab:osm_categories} in the appendix.
\subsection{Image Modality}
\par The image modality is constructed from Sentinel-2 Level-2A imagery. Level-2A products provide atmospherically corrected surface reflectance and include the {$10$} m RGB bands used in this work \cite{copernicusSentinel2Products}. For each H3 anchor, we retrieve a fixed-size crop of $512 \times 512$ pixels,
i.e., $5.12 \times 5.12$ km at the native 10 m resolution, centered on the
cell centroid, so that the crop always contains the full target cell together
with its surroundings (at resolution~9, the parent-cell crop is
reused). We use only the visible RGB bands B4, B3, and B2 \eqref{eq4}:
\begin{equation}
    v_i = [\mathrm{B4}_i, \mathrm{B3}_i, \mathrm{B2}_i],
    \label{eq4}
\end{equation}
where B2 corresponds to blue, B3 to green, and B4 to red reflectance, each at {$10$} m spatial resolution in Sentinel-2 MSI products \cite{copernicusSentinel2Products}.
\par Candidate scenes are selected from the March 2025--February 2026 acquisition window, with preference for observations closest to December 2025, matching the temporal reference of the OSM snapshot. We prioritize cloud-free observations and treat {$20\%$} cloud coverage as an upper bound. Approximately {$5\%$} of cells require an observation outside the primary window because no scene satisfying these constraints was available within it.
\par If an H3 crop overlaps multiple Sentinel-2 tiles or scene boundaries, we construct the image by mosaicking the available scene fragments. To avoid incomplete or visually truncated crops, we apply a simple histogram-based validation procedure that detects abrupt missing regions and abnormal intensity distributions. Crops failing this validation are replaced by the next-best candidate scene according to cloud coverage and temporal distance from December 2025. Digital values are converted to reflectance using the Sentinel-2 scale factor and then normalized with the channel statistics of the frozen visual encoder.
\subsubsection{Hexagon-Aware Patch-Token Pooling}
\label{subsec:patch_pooling}
\par The image view is produced by a frozen DINOv3 ViT-L/16 backbone, using the satellite-pretrained \texttt{sat493m} variant of the released checkpoint \cite{simeoni2025dinov3}. Each validated crop is used at its native $512 \times 512$ size, which the
16-pixel patch embedding turns into a regular $32 \times 32$ grid of 1024 patch
tokens, each covering $160 \times 160$\,m on the ground. The way this grid is reduced to a single vector is the one place where our image branch deviates substantially from standard practice, and the deviation is motivated by the geometry of the H3 grid rather than by encoder design.
\par The natural choice, the \texttt{CLS} token, is a global descriptor of the entire square crop. In our setting the square crop necessarily extends beyond the target H3 hexagon, so a global descriptor mixes the target cell with surrounding context and is less spatially specific than the H3-indexed graph and text views. We therefore use patch tokens to construct a representation whose receptive field is matched more closely to the footprint of the target hexagon.
\par Concretely, we pool only a central window of patch tokens. Let $t_s$ denote the token at grid position $s$ and let $\mathcal{W}_r(h_i)$ be the window selected for cell $h_i$ at H3 resolution $r$. The stored image embedding $e_i^v$ is the L2-normalized mean of that window \eqref{eq:patchpool}:
\begin{equation}
    e_i^v
    = \frac{\bar{t}_i}{\|\bar{t}_i\|_2},
    \qquad
    \bar{t}_i
    = \frac{1}{|\mathcal{W}_r(h_i)|}
      \sum_{s \in \mathcal{W}_r(h_i)} t_s .
    \label{eq:patchpool}
\end{equation}
\par Because all crops share the same ground extent while H3 cells do not, the
window side length is resolution-dependent: we use $14 \times 14$, $5 \times 5$,
and $2 \times 2$ token windows for resolutions 7, 8, and 9. These sizes are
chosen so that the area of the pooled window matches the average cell area, which keeps the effective receptive field of the image view aligned with the spatial unit that the graph and text views describe. Tokens are read before the final layer normalization of the backbone, and normalization is applied once, after pooling, so that the pooled vector lies on the unit sphere used by the contrastive objective.
\par At resolution {$9$} the same mechanism yields a further benefit. A resolution-8 parent cell has seven resolution-9 children, and all seven fall inside the parent crop. Instead of encoding seven crops, we encode the parent once and read seven disjoint {$2 \times 2$} windows from the resulting token grid: one at the center and six arranged around it. The assignment is geometric rather than arbitrary. We sort the seven children by their azimuth relative to the parent centroid and match that ordering to the azimuths of the six peripheral windows, so each window is registered to the child hexagon whose ground area it covers. A single forward pass therefore produces seven spatially registered child embeddings, which reduces encoder cost at the finest resolution by roughly a factor of seven and guarantees that sibling cells are described in a consistent visual context.
\subsection{Text Modality}
\par The text modality supplies place semantics that geometry alone does not expose and combines descriptions associated with OSM-derived objects with three complementary external sources. Wikipedia contributes encyclopedic descriptions of georeferenced places, Wikivoyage contributes travel-oriented accounts of urban areas, and Reddit contributes informal local discourse from city-specific communities.
\par For each metropolitan area we query a regular grid of points with a {$4$} km step and a {$5$} km search radius, so that the overlapping circles cover large regions without relying on a single city-level query; Wikipedia and Wikivoyage results are then deduplicated by page identifier. From Wikipedia we keep the introductory plain-text summary and discard very short extracts. From Wikivoyage we keep the initial descriptive part of each guide, truncated before operational sections such as ``Get in'' or ``Get around'', which retains the description of a place rather than lists of attractions and transport instructions. Reddit posts are usually not geotagged, so grid points are reverse-geocoded to city and neighborhood names, the corresponding city subreddit is queried by neighborhood name with a looser keyword fallback, and post titles with truncated self-text are retained. This source is noisier but captures colloquial place names, safety perceptions, nightlife, and housing discussion.
\par External records are mapped to H3 by coordinates, while OSM-associated descriptions are keyed to the corresponding H3 anchor; texts assigned to the same cell are concatenated into one cell-level document. Resolution-7 coverage comes from cell-level OSM-associated or external descriptions, while finer cells without direct text inherit the parent document. External content was accessed in March 2026, whereas the OSM-associated component follows the December 2025 object snapshot.
\par The encoder itself is used off the shelf: cleaned cell-level documents are chunked when they exceed the maximum sequence length, each chunk is encoded independently by the frozen \texttt{jinaai/\allowbreak jina-embeddings-v3} multilingual model \cite{sturua2025jina}, chunk embeddings of the same cell are averaged, and the result is reduced to a {$128$}-dimensional Matryoshka representation and L2-normalized \eqref{eq5}:
\begin{equation}
t_i = f_{\mathrm{text}}(D_i),
\label{eq5}
\end{equation}
where {$D_i$} is the aggregated document for cell {$h_i$} and {$f_{\mathrm{text}}$} denotes the frozen encoder with the preprocessing above. The dataset release reports per-source page counts, cell coverage, the share of propagated versus directly assigned text, token statistics, and the language distribution.
\subsection{Pairing and Temporal Consistency}
\par The central design principle of the dataset is H3-based pairing. For every retained anchor {$h_i$}, the graph, image, and text views refer to the same geographic region. This design avoids approximate joins between incompatible spatial units and allows contrastive learning to treat {$(g_i, v_i, t_i)$} as positive views of the same location.
\par The three modalities are not perfectly simultaneous, which is unavoidable when combining map data, satellite imagery, and web text. We therefore define December 2025 as the primary temporal reference point. The graph modality uses the December 2025 OSM snapshot. The image modality uses Sentinel-2 observations from March 2025 to February 2026, preferring observations closest to December 2025 and allowing exceptions only when no valid low-cloud scene is available. The text modality uses the December 2025 OSM object set but reflects the state of external web sources accessed in March 2026. We report these temporal choices explicitly to avoid implying exact simultaneity across modalities.
\section{GeoLEON-CLIP}
\label{sec:model}
%
\subsection{Architecture}
\par GeoLEON-CLIP maps graph, text, and image representations into a shared contrastive space. Let {$e_i^g$}, {$e_i^t$}, and {$e_i^v$} denote frozen inputs for H3 anchor {$h_i$}. The graph and text branches concatenate parent, center, and pooled child embeddings, whereas the image branch uses the center-cell embedding from the patch-token pooling of Section~\ref{subsec:patch_pooling}. The projected representation is \eqref{eq6}:
\begingroup
\setlength{\abovedisplayskip}{3pt}
\setlength{\belowdisplayskip}{3pt}
\setlength{\abovedisplayshortskip}{3pt}
\setlength{\belowdisplayshortskip}{3pt}
\begin{equation}
    z_i^m = \frac{p_m(e_i^m)}{\|p_m(e_i^m)\|_2}, \quad m \in \{g,t,v\},
    \label{eq6}
\end{equation}
\endgroup
where {$p_m$} is a trainable modality-specific projection head. The best-performing model uses residual MLP heads with hidden dimension {$4096$}, three residual layers, GELU activations, layer normalization, and dropout of {$0.1$}. The shared projection dimension is {$256$}. Encoders are frozen; only the projection heads and logit scale are optimized.
\par The projection-head design is deliberately lightweight relative to full encoder pretraining. This lets the experiment test whether already learned graph, image, and text representations can be geometrically aligned through a small trainable layer. Dropped modalities are represented by zeros in the base model; additional runs evaluate learned missing tokens.
\subsection{Tri-Modal Contrastive Objective}
\par For a minibatch of {$N$} H3 anchors, GeoLEON-CLIP optimizes all pairwise modality alignments. For modality pair {$(a,b)$}, similarities are computed as \eqref{eq7}:
\begin{equation}
    s_{ij}^{ab} = \tau \, (z_i^a)^\top z_j^b,
     \label{eq7}
\end{equation}
where {$\tau$} is a learned logit scale clamped to a maximum value of {$100$}. The directional contrastive loss is \eqref{eq8}:
\begin{equation}
    \mathcal{L}_{a \rightarrow b}
    = - \frac{1}{N}\sum_i
    \log
    \frac{\exp(s_{ii}^{ab})}
    {\sum_j \exp(s_{ij}^{ab})}.
     \label{eq8}
\end{equation}
\par The total objective is the sum of the three symmetric pairwise losses for the
graph--text, graph--image, and text--image pairs, each averaging its two
directional terms \eqref{eq9}:
\begin{equation}
    \mathcal{L}
    =
    \mathcal{L}_{g,t}
    + \mathcal{L}_{g,v}
    + \mathcal{L}_{t,v},
    \quad
    \mathcal{L}_{a,b}
    =
    \frac{1}{2}(\mathcal{L}_{a \rightarrow b}+\mathcal{L}_{b \rightarrow a}).
     \label{eq9}
\end{equation}
\par The three pair losses have equal weight: the objective does not privilege graph-involved pairs. ``Graph-anchored'' instead describes the use of the pretrained LEON representation and the graph branch's empirically observed bridge behavior. This role emerged from uniform tri-modal alignment rather than from an asymmetric loss designed to favor it. The inputs are nevertheless asymmetric in spatial context: LEON graph embeddings already incorporate neighborhood message passing and are further combined with parent/center/child context, the text branch uses the hierarchical parent/center/child context, and the image branch uses only the center-cell visual embedding. This asymmetry is a plausible contributor to the graph's bridge role and to the weaker image--text alignment; extending the image branch with parent/child visual context is a natural direction for future work.
\par The main model uses diagonal positives only. Additional ablations evaluate cross-resolution multi-positive masks, hard-negative upweighting, modality dropout, learned missing tokens, and smaller or linear projection heads.
\subsection{Training Setup}
\par The reported base run trains for {$10$} epochs with batch size 1024, AdamW optimization, learning rate {$10^{-4}$} for projection heads, logit-scale learning rate {$10^{-2}$}, {$500$} warmup steps, cosine decay, gradient clipping at {$1.0$}, and exponential moving average weights for evaluation. During training, graph, text, and image are independently dropped with probabilities 0.10, 0.05, and 0.10; dropout is disabled at evaluation. We split 975,432 anchors once, with a fixed seed, into 877,889 training (90\%) and 97,543 validation (10\%) anchors, pooling all cities and resolutions and applying no spatial blocking. LEON was pretrained self-supervised on graphs from all 35 cities, without image or text supervision. All pairwise and chained retrieval results use one fixed 1024-anchor validation query/gallery pool and the same EMA checkpoint. Because the pool is sampled proportionally across the pooled resolutions, it is dominated by resolution 9 (about 85\% of the sampled anchors). Ablation results reported below use the same validation protocol as the base checkpoint.
\section{Evaluation Protocol}
\label{sec:evaluation}
\par Evaluation covers pairwise and chained retrieval, distance-aware geographic error, embedding geometry, and downstream prediction probes.
\paragraph{Cross-modal retrieval.}
The primary evaluation is retrieval between each ordered pair of modalities: graph {$\rightarrow$} text, text {$\rightarrow$} graph, graph {$\rightarrow$} image, image {$\rightarrow$} graph, text {$\rightarrow$} image, and image {$\rightarrow$} text. We report Recall@1, Recall@5, Recall@10, median rank, and mean reciprocal rank (MRR). These metrics directly test whether paired H3 views are close in the learned space.
\par Because geographic retrieval errors have spatial structure, we additionally report a non-standard distance-aware metric, mean minimum geographic error (MGE). MGE measures how far the best geographically nearby retrieved candidate is from the true H3 cell, rather than only whether the exact cell appears at a given rank. For each query and cutoff {$k$}, MGE@{$k$} computes the haversine distance, in kilometers, between each of the top-{$k$} retrieved H3 cell centroids and the positive target cell centroid, keeps the minimum distance among those {$k$} candidates, and averages this value over valid queries \eqref{eq10}:
\begin{equation}
    \mathrm{MGE@}k =
    \frac{1}{|\mathcal{Q}|}
    \sum_{i \in \mathcal{Q}}
    \min_{j \in \mathrm{Top}_k(i)}
    d_{\mathrm{hav}}(h_j, h_i^+).
    \label{eq10}
\end{equation}
\par For multi-positive evaluation, {$h_i^+$} is the closest valid positive target. Lower MGE is better. This metric complements rank-based retrieval because an incorrect match in a nearby district is less geographically severe than an incorrect match on another continent.
\paragraph{Chained cross-modal recall.}
\par Pairwise retrieval establishes that the graph aligns well with both other modalities, but it does not test the anchoring claim itself. The claim is that the graph is a useful \emph{intermediate} representation, and the direct way to test it is to force retrieval to pass through the graph and ask whether the detour costs anything. We therefore define a two-hop metric. Let $a$ be the query modality, $b$ the bridge modality, and $c$ the target modality. For a query cell $h_i$, the first hop retrieves the top-$A$ cells in the bridge modality, and the second hop retrieves the top-$B$ target cells for each of them, giving the chained candidate set \eqref{eq:chainset}:
\begin{equation}
    \mathcal{C}_{A,B}(i)
    = \bigcup_{j \in \mathrm{Top}_A^{a \rightarrow b}(i)}
      \mathrm{Top}_B^{b \rightarrow c}(j),
    \label{eq:chainset}
\end{equation}
and chained recall is the fraction of queries whose target is contained in that set \eqref{eq:chainrecall}:
\begin{equation}
    \mathrm{R}@A@B
    = \frac{1}{|\mathcal{Q}|}
      \sum_{i \in \mathcal{Q}}
      \mathbf{1}\!\left[h_i^+ \in \mathcal{C}_{A,B}(i)\right].
    \label{eq:chainrecall}
\end{equation}
\par The natural reference point is direct $a \rightarrow c$ retrieval at a matched candidate budget. Since $|\mathcal{C}_{A,B}(i)| \leq AB$, with equality only when the second hops return disjoint results, we compare $\mathrm{R}@A@B$ against the direct $\mathrm{R}@(AB)$. This comparison is deliberately conservative: whenever the second hops overlap, the chain inspects strictly fewer distinct candidates than the baseline it is measured against. The single exception is $A=B=1$, where the budget is exactly one candidate on both sides and no deduplication slack exists. $\mathrm{R}@1@1$ is therefore the strict form of the metric, and it is also the harshest one, because the chain has to commit to a single intermediate cell with no opportunity to recover from a first-hop error.
\paragraph{Embedding geometry.}
We measure alignment by the average cosine similarity of paired normalized embeddings and characterize isotropy through the effective dimension of the projected embedding covariance spectrum. If $\lambda_j$ are the covariance eigenvalues, $d_{\mathrm{eff}}=(\sum_j\lambda_j)^2/(\sum_j\lambda_j^2)$; larger values indicate that variance is distributed across more directions. We report effective dimension for each modality and for the pooled joint space.
\paragraph{Downstream probes.}
We also evaluate whether pre-projection and post-projection embeddings improve supervised prediction when appended to task-specific baseline features. The probe model is Ridge regression. We evaluate five regression tasks: Airbnb prices, Beijing housing, Chicago crime, King County house sales, and Philadelphia crime. For each task, we compare a no-embedding baseline against graph, text, image, and all-modality embeddings, both before and after projection, and report the best-performing configuration per task in terms of {$R^2$}.

\section{Results}
\label{sec:results}
\par The base residual-head model achieves the strongest cross-modal retrieval performance on the graph--image pair, followed by graph--text, while text--image retrieval is weaker but still meaningful.
\subsection{Cross-Modal Retrieval}
\par Table~\ref{tab:retrieval_base} reports retrieval performance for the base residual-head model. Graph--image is the strongest pair, reaching Recall@1 of {$0.800$} and {$0.813$}; graph--text follows at {$0.742$} and {$0.741$}. Direct text--image retrieval remains meaningful at Recall@1 of 0.616--0.617 and Recall@10 above {$0.84$} in both directions.
\begin{table}[t]
\centering
\caption{Cross-modal retrieval for the base GeoLEON-CLIP model. The base run uses residual MLP projection heads and 1024 validation queries.}
\label{tab:retrieval_base}
\small
\setlength{\tabcolsep}{4pt}
\renewcommand{\arraystretch}{1.12}
\begin{tabular}{@{}l r r r r r@{}}
\toprule
\textbf{Direction} & \textbf{R@1} & \textbf{R@5} & \textbf{R@10} & \textbf{Med.} & \textbf{MRR} \tabularnewline
\midrule
Graph $\rightarrow$ Text & 0.742 & 0.884 & 0.930 & 1 & 0.806 \tabularnewline
Text $\rightarrow$ Graph & 0.741 & 0.878 & 0.928 & 1 & 0.804 \tabularnewline
Graph $\rightarrow$ Image & 0.800 & 0.924 & 0.958 & 1 & 0.859 \tabularnewline
Image $\rightarrow$ Graph & 0.813 & 0.932 & 0.969 & 1 & 0.869 \tabularnewline
Image $\rightarrow$ Text & 0.617 & 0.778 & 0.847 & 1 & 0.696 \tabularnewline
Text $\rightarrow$ Image & 0.616 & 0.783 & 0.846 & 1 & 0.697 \tabularnewline
\bottomrule
\end{tabular}
\end{table}
\par Table~\ref{tab:mge_base} reports the corresponding MGE values. Image-to-graph gives the strongest geographic localization, reaching MGE@10 of {$37.7$} km, followed by graph-to-image at {$78.5$} km and the two graph--text directions at 103.0--115.3 km. Text-to-image remains the hardest direction at {$325.9$} km.
\begin{table}[t]
\centering
\caption{Mean minimum geographic error (MGE) for the base GeoLEON-CLIP model. Values are kilometers; lower is better. MGE@$k$ uses the closest H3 centroid among the top-$k$ retrieved candidates.}
\label{tab:mge_base}
\small
\setlength{\tabcolsep}{5pt}
\renewcommand{\arraystretch}{1.12}
\begin{tabular}{@{}l r r r@{}}
\toprule
\textbf{Direction} & \textbf{MGE@1} & \textbf{MGE@5} & \textbf{MGE@10} \tabularnewline
\midrule
Graph $\rightarrow$ Text & 1541.8 & 300.9 & 115.3 \tabularnewline
Text $\rightarrow$ Graph & 1625.5 & 250.2 & 103.0 \tabularnewline
Graph $\rightarrow$ Image & 819.4 & 185.3 & 78.5 \tabularnewline
Image $\rightarrow$ Graph & 790.2 & 74.9 & 37.7 \tabularnewline
Image $\rightarrow$ Text & 1781.1 & 314.8 & 127.7 \tabularnewline
Text $\rightarrow$ Image & 1931.6 & 541.9 & 325.9 \tabularnewline
\bottomrule
\end{tabular}
\end{table}
\par These results are consistent with the graph-anchor hypothesis, which we test
directly in Section~\ref{subsec:chained_recall}. In the mixed-resolution pool, graph--image is strongest, graph--text follows, and direct text--image alignment is hardest. The ordering is not scale-invariant: in a separate resolution-8 gallery, G$\leftrightarrow$T reaches 0.86/0.87 R@1 versus 0.78/0.82 for G$\leftrightarrow$I, while text--image remains the weakest pair at every level. A plausible contributor at the finest scale is hierarchical text propagation: some resolution-9 cells inherit a parent description, whereas graph and image features remain cell-local.
\par As a limited zero-shot sanity check on the same resolution-8 cells, CLIP, RemoteCLIP, and GeoRSCLIP reach at most 1.5\% R@1 and 11.5\% R@10 for I$\leftrightarrow$T, versus 80.8\% R@1 for GeoLEON-CLIP direct retrieval (92.3\%/91.5\% with graph-routed $1@1$). This is not a like-for-like benchmark. As a partial encoder control, we retrained identical projection heads on a 10\% subset for 850 steps (one seed, no EMA), changing only the frozen image features to RemoteCLIP-B/32 or GeoRSCLIP-B/32. Relative to DINOv3, G$\leftrightarrow$I R@1 fell by 0.045--0.057 and I$\leftrightarrow$T by 0.015--0.031; the 95\% bootstrap interval of the RemoteCLIP I$\to$T difference included zero. Thus encoder choice alone does not explain the zero-shot gap, but this reduced experiment is only a partial baseline comparison.

\providecommand{\RetrievalImage}{}
\renewcommand{\RetrievalImage}[1]{%
    \begin{minipage}[c][3.15cm][c]{0.40\textwidth}%
        \centering
        \includegraphics[width=0.92\linewidth,height=3.05cm,keepaspectratio]{#1}%
    \end{minipage}%
}

\providecommand{\RetrievalGraph}{}
\renewcommand{\RetrievalGraph}[1]{%
    \begin{minipage}[c][3.15cm][c]{0.40\textwidth}%
        \centering
        \includegraphics[width=0.92\linewidth,height=3.05cm,keepaspectratio]{#1}%
    \end{minipage}%
}

\providecommand{\RetrievalText}{}
\renewcommand{\RetrievalText}[1]{%
    \begin{minipage}[c][3.15cm][c]{0.40\textwidth}%
        \raggedright
        \small\itshape
        ``#1''%
    \end{minipage}%
}

\providecommand{\RetrievalArrow}{}
\renewcommand{\RetrievalArrow}{%
    \begin{minipage}[c][3.15cm][c]{0.08\textwidth}%
        \centering
        \Large$\Longrightarrow$%
    \end{minipage}%
}

\par
\begin{figure*}[t]
    \centering
    \setlength{\tabcolsep}{0pt}
    \renewcommand{\arraystretch}{1.0}

    \begin{tabular}{@{}c@{\hspace{0.015\textwidth}}c@{\hspace{0.015\textwidth}}c@{}}
        \begin{minipage}[c]{0.40\textwidth}\centering\textbf{Query: text}\end{minipage}
        &
        \begin{minipage}[c]{0.08\textwidth}\centering\end{minipage}
        &
        \begin{minipage}[c]{0.40\textwidth}\centering\textbf{Top-1 retrieved satellite image}\end{minipage}
        \tabularnewline[0.25em]

        \RetrievalText{A transit-oriented district centered on Yatsuka Station, characterized by railway access and everyday mobility functions that shape local movement and connectivity.}
        &
        \RetrievalArrow
        &
        \RetrievalImage{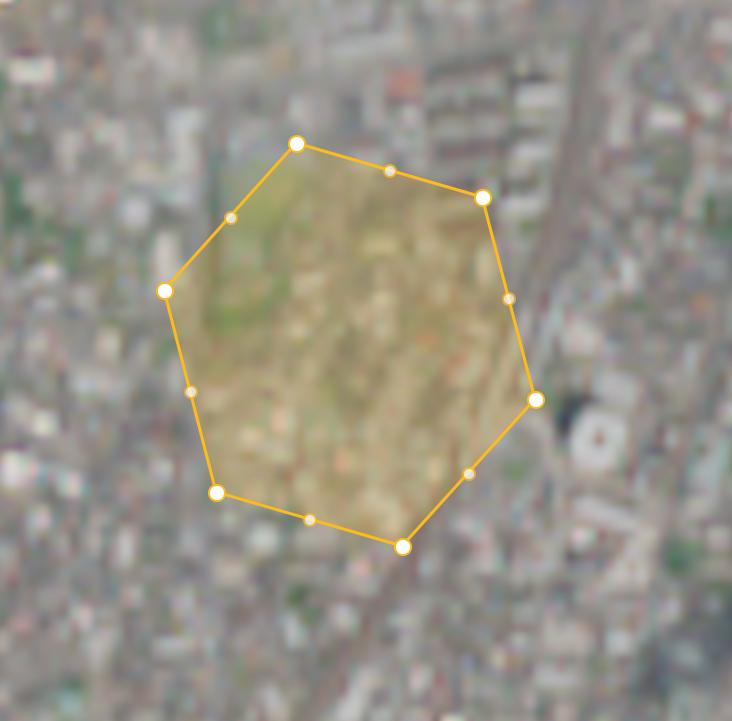}
        \tabularnewline[0.65em]

        \begin{minipage}[c]{0.40\textwidth}\centering\textbf{Query: graph region}\end{minipage}
        &
        \begin{minipage}[c]{0.08\textwidth}\centering\end{minipage}
        &
        \begin{minipage}[c]{0.40\textwidth}\centering\textbf{Top-1 retrieved text}\end{minipage}
        \tabularnewline[0.25em]

        \RetrievalGraph{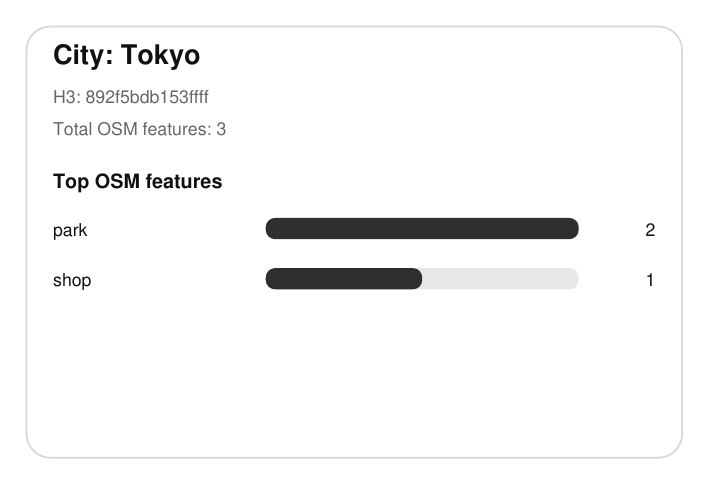}
        &
        \RetrievalArrow
        &
        \RetrievalText{A calm residential district with sparse mapped amenities, defined mainly by green space, a playground, a worship facility, and basic retail services supporting local community life.}
        \tabularnewline[0.65em]

        \begin{minipage}[c]{0.40\textwidth}\centering\textbf{Query: satellite image}\end{minipage}
        &
        \begin{minipage}[c]{0.08\textwidth}\centering\end{minipage}
        &
        \begin{minipage}[c]{0.40\textwidth}\centering\textbf{Top-1 retrieved graph region}\end{minipage}
        \tabularnewline[0.25em]

        \RetrievalImage{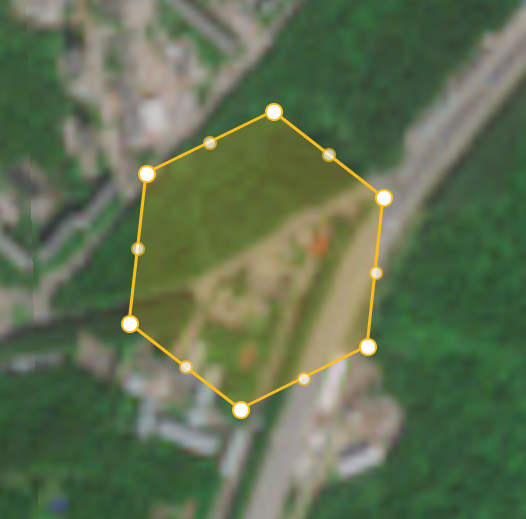}
        &
        \RetrievalArrow
        &
        \RetrievalGraph{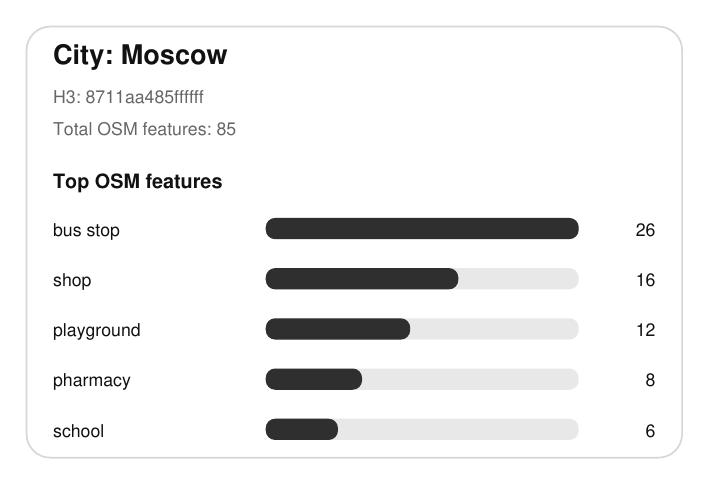}
        \tabularnewline
    \end{tabular}

    \caption{Qualitative cross-modal retrieval examples involving the graph modality. Each row shows a query from one modality and the top-1 retrieved result from another modality. Graph regions are represented by H3/OSM graph cards, satellite regions by image crops, and text by short semantic descriptions.}
    \label{fig:qualitative-retrieval-graph}
\end{figure*}

 Figure \ref{fig:qualitative-retrieval-graph} presents qualitative cross-modal retrieval examples involving the graph modality, where each row displays a query from one modality (graph, satellite image, or text) and its top-1 retrieved result from another modality. Graph regions are visualized as H3/OSM graph cards, satellite regions as image crops, and text as short semantic descriptions.
\subsection{Graph as a Retrieval Bridge}
\label{subsec:chained_recall}
\par Table~\ref{tab:chained_recall} evaluates both graph-bridged directions across selected budgets and, at $1@1$, all four alternative bridge placements. Chained and direct values use the same checkpoint, 1024-cell pool, and evaluator as Table~\ref{tab:retrieval_base}; the direct column is therefore exactly the corresponding endpoint retrieval at $k=AB$.
\begin{table}[t]
\centering
\caption{Chained recall and direct endpoint retrieval at matched budget $k=AB$ (\%). G, I, and T denote graph, image, and text. Alternative bridges are reported at $1@1$; $\Delta$ is computed from exact hit counts before percentage rounding.}
\label{tab:chained_recall}
\scriptsize
\setlength{\tabcolsep}{3pt}
\renewcommand{\arraystretch}{1.06}
\begin{tabular}{@{}l r r r r@{}}
\toprule
\textbf{Chain} & $A@B$ & \textbf{Chained} & \textbf{Direct} & $\Delta$ \tabularnewline
\midrule
I$\rightarrow$G$\rightarrow$T & 1@1 & \textbf{66.5} & 61.7 & \textbf{+4.8} \tabularnewline
 & 1@5 & 78.6 & 77.8 & +0.8 \tabularnewline
 & 3@3 & 81.5 & 83.9 & $-$2.3 \tabularnewline
 & 5@5 & 87.6 & 94.1 & $-$6.5 \tabularnewline
 & 10@1 & 74.9 & 84.7 & $-$9.8 \tabularnewline
\addlinespace[1pt]
T$\rightarrow$G$\rightarrow$I & 1@1 & \textbf{66.0} & 61.6 & \textbf{+4.4} \tabularnewline
 & 1@5 & 75.8 & 78.3 & $-$2.5 \tabularnewline
 & 3@3 & 81.1 & 83.8 & $-$2.7 \tabularnewline
 & 5@5 & 87.9 & 91.8 & $-$3.9 \tabularnewline
 & 10@1 & 77.4 & 84.6 & $-$7.1 \tabularnewline
\midrule
G$\rightarrow$T$\rightarrow$I & 1@1 & 59.6 & 80.0 & $-$20.4 \tabularnewline
I$\rightarrow$T$\rightarrow$G & 1@1 & 60.1 & 81.3 & $-$21.3 \tabularnewline
G$\rightarrow$I$\rightarrow$T & 1@1 & 58.4 & 74.2 & $-$15.8 \tabularnewline
T$\rightarrow$I$\rightarrow$G & 1@1 & 57.6 & 74.1 & $-$16.5 \tabularnewline
\bottomrule
\end{tabular}
\end{table}
\par The strict setting is the clearest result. At $\mathrm{R}@1@1$, I$\rightarrow$G$\rightarrow$T beats direct retrieval by 4.8 points (66.5\% versus 61.7\%), while T$\rightarrow$G$\rightarrow$I gains 4.4 points (66.0\% versus 61.6\%); paired-bootstrap 95\% intervals for the gains are [2.3, 7.3] and [1.8, 7.1] points. Each chain makes two top-1 decisions and cannot recover from a first-hop error, so these gains directly quantify the graph's high-precision bridge role.
\par The advantage largely disappears as the budget grows: only I$\rightarrow$G$\rightarrow$T retains a small +0.8 point margin at $1@5$, and direct retrieval wins the other selected wider-budget comparisons. Evaluating each resolution separately at $1@1$ gives positive graph-route point estimates at every level, with observed margins of about 10--13 points at resolutions 7--8 and 2--4 points at resolution 9; the T$\rightarrow$G$\rightarrow$I interval at resolution 9 includes zero, and all four non-graph bridges remain negative. These values show scale dependence, but differing gallery difficulty prevents interpreting them as a ranking of H3 levels. Full budget grids are given in Appendix~\ref{app:chained-grids}.
\subsection{Projection Head and Training Ablations}
\par Table~\ref{tab:ablation_retrieval} compares the base checkpoint with projection-head and training ablations under the same evaluation protocol. The residual head is strongest in all six directions, and linear projection degrades every pair. Removing modality dropout is also costly for image-involved retrieval (G$\rightarrow$I 0.585 vs. 0.800; I$\rightarrow$T 0.366 vs. 0.617; T$\rightarrow$I 0.350 vs. 0.616), making modality dropout the second clearest contributor. Multi-positive, hard-negative, and learned-token variants remain closer to the base, while the small MLP remains below it in every direction.
\begin{table*}[t]
\centering
\caption{Recall@1 for the base GeoLEON-CLIP checkpoint and ablations evaluated under the same protocol.}
\label{tab:ablation_retrieval}
\scriptsize
\setlength{\tabcolsep}{4pt}
\renewcommand{\arraystretch}{1.12}
\begin{tabular}{@{}l r r r r r r@{}}
\toprule
\textbf{Variant} & \textbf{G$\rightarrow$T} & \textbf{T$\rightarrow$G} & \textbf{G$\rightarrow$I} & \textbf{I$\rightarrow$G} & \textbf{I$\rightarrow$T} & \textbf{T$\rightarrow$I} \tabularnewline
\midrule
Base residual head & \textbf{0.742} & \textbf{0.741} & \textbf{0.800} & \textbf{0.813} & \textbf{0.617} & \textbf{0.616} \tabularnewline
Linear head & 0.523 & 0.526 & 0.295 & 0.295 & 0.258 & 0.249 \tabularnewline

Small MLP & 0.615 & 0.592 & 0.400 & 0.413 & 0.297 & 0.304 \tabularnewline
No modality dropout & 0.716 & 0.710 & 0.585 & 0.611 & 0.366 & 0.350 \tabularnewline
Multi-positive & 0.703 & 0.700 & 0.771 & 0.794 & 0.526 & 0.505 \tabularnewline
Hard negatives & 0.701 & 0.701 & 0.768 & 0.783 & 0.523 & 0.515 \tabularnewline
Learned tokens & 0.706 & 0.701 & 0.772 & 0.786 & 0.532 & 0.514 \tabularnewline
\bottomrule
\end{tabular}
\end{table*}
\subsection{Embedding Geometry}
\par The projected space shows non-collapsed geometry. Post-projection effective dimensions are {$28.04$} for graph, {$22.93$} for text, {$24.05$} for image, and {$29.62$} when all modalities are pooled. Average cross-modal cosine alignment is {$0.571$} for text--graph, {$0.442$} for image--graph, and {$0.302$} for text--image.
\subsection{Downstream Prediction}
\par Table~\ref{tab:downstream_best} summarizes the best embedding-augmented configuration for each downstream task. The improvements are strongest for Airbnb and Beijing housing. Airbnb improves from {$R^2=0.216$} to {$0.287$} using post-projection all-modality features, a gain of 0.071. Beijing housing improves from {$0.669$} to {$0.690$} using pre-projection all-modality features. Smaller positive gains appear for King County and Philadelphia crime, while Chicago crime has an extremely high baseline and no measurable gain from pre-projection graph features. The downstream prediction tasks come from \cite{moska2025obsr}.
\begin{table}[t]
\centering
\caption{Best downstream Ridge probe result per task when embeddings are appended to baseline task features.}
\label{tab:downstream_best}
\small
\setlength{\tabcolsep}{4pt}
\renewcommand{\arraystretch}{1.12}
\begin{tabular}{@{}l l r r r@{}}
\toprule
\textbf{Task} & \textbf{Best config.} & \textbf{Base $R^2$} & \textbf{Best $R^2$} & $\Delta$ \tabularnewline
\midrule
Airbnb & Post / all & 0.216 & 0.287 & +0.071 \tabularnewline
Beijing housing & Pre / all & 0.669 & 0.690 & +0.021 \tabularnewline
Chicago crime & Pre / graph & 0.978 & 0.978 & +0.000 \tabularnewline
King County & Post / image & 0.762 & 0.764 & +0.002 \tabularnewline
Philadelphia crime & Post / image & 0.876 & 0.883 & +0.007 \tabularnewline
\bottomrule
\end{tabular}
\end{table}
\par The downstream results should be interpreted as evidence of useful representation signal, not as universal dominance over strong tabular baselines. Several all-modality configurations degrade on high-baseline crime and housing tasks. The positive pattern is more specific: embeddings help most when the baseline leaves room for visual, semantic, or neighborhood context, and the best modality mixture varies by task.
\subsection{Exploratory Cross-Modal Embedding Arithmetic}
\par We also ran an exploratory qualitative analysis of arithmetic in the aligned space, motivated by earlier work on spatial embedding arithmetic such as Hex2Vec \cite{wozniak2021hex2vec} but differing in two respects: the arithmetic is performed in a tri-modal aligned space rather than a purely OSM-derived one, and the result is evaluated by retrieval in a third modality. We consider expressions of the form
\[
    z_i^a + z_j^b \rightarrow \operatorname{retrieve}(z^c),
    \qquad a,b,c \in \{g,v,t\}, \; a,b,c \text{ distinct},
\]
so that, for example, an image embedding from one H3 cell is combined with a text embedding from another and the nearest graph embeddings are retrieved. The aim is not a quantified compositional benchmark but an inspection of whether the space behaves compositionally at all.
\par The arithmetic does not appear to act only on the explicitly selected modality. Although each expression uses one modality per input region, the retrieved result often reflects properties of the whole tri-modal object associated with that region, indicating that a single-modality embedding still carries information induced by its alignment with the other two views of the same cell. This is consistent with the objective, in which each projected representation is trained as one view of a shared region representation rather than as an isolated modality-specific code.
\par The two operands also tend to play different roles. One contributes the functional or semantic character of a place, particularly when the graph modality is involved, which is plausible given that the graph branch encodes OSM categories and spatial adjacency. The other tends to influence geographic placement, with retrieved regions frequently staying within or near the metropolitan area of one operand. Notably, the text operand alone was sometimes sufficient to bias placement toward the corresponding city, despite being the modality least directly tied to spatial geometry, suggesting that place descriptions carry implicit geographic cues that become spatially grounded through contrastive alignment.
\par These observations are preliminary and qualitative; counterexamples occur, and a complete evaluation would require controlled arithmetic benchmarks, explicit separation of functional similarity from geographic proximity, and statistical tests over many city and modality combinations. Examples are provided in Appendix~\ref{app:cross-modal-arithmetic}.

\section{Discussion and Conclusion}
\label{sec:discussion}
\par The retrieval experiments provide the clearest evidence for GeoLEON-CLIP. In the mixed-resolution pool, graph--image is strongest, graph--text follows, and text--image is weakest, but the per-resolution check shows that the first two can reverse order. The robust pattern is therefore that graph aligns strongly with both endpoints while direct image--text retrieval is consistently harder, supporting its use as an intermediate representation.
\par Chained recall turns that ordering into a stronger statement. A strictly single-candidate detour through graph is more accurate than direct image--text matching in both directions, while the alternative bridge placements that traverse the weaker image--text relation are substantially worse than direct endpoint retrieval. The $\mathrm{R}@1@1$ result therefore suggests that the advantage occurs when the intermediate modality is strongly aligned with both endpoints. Its limited budget range is equally informative: as the candidate set grows, direct retrieval generally regains the advantage, so graph should be read as a high-precision bridge rather than a general replacement for direct retrieval.
\par The downstream probes are more nuanced. Adding embeddings improves several tasks, especially Airbnb and Beijing housing, but no single modality configuration dominates across all datasets: some tasks already have strong baseline features, and concatenating all modalities can introduce noise or redundant signal. The aligned embeddings therefore contain transferable geographic signal, but exploiting it in prediction tasks requires more care than concatenation. Future work should evaluate task-adaptive fusion, stronger spatial splits, multi-seed downstream testing, and end-to-end fine-tuning of selected encoders.
\par Taken together, the results support graph-anchored multimodal learning as a direction for geospatial representation learning: an OSM-derived graph representation, trained without labels and kept frozen, can help organize satellite imagery and place descriptions in a shared H3-indexed space, with graph routing acting as a high-precision bridge whose observed margin varies with H3 resolution.
\subsection{Study Limitations}
\par The dataset has several limitations. OSM coverage is geographically and socially uneven, textual sources vary across cities and languages, and the three modalities are only approximately aligned in time. The model is trained on precomputed image embeddings, so acquisition dates,
cloud coverage, and mosaicking provenance are not exposed to it, and their
effect on alignment is not analyzed. Text is also partly inherited across the H3 hierarchy: finer cells without directly assigned text use their parent-cell document, so sibling cells may share the same text view.
\par The principal evaluation limitation is spatial generalization. The random anchor-level split contains all cities and resolutions in both partitions and uses no spatial blocking; together with graph-only LEON pretraining on all 35 cities, it can benefit from spatial autocorrelation and does not measure transfer to an unseen city. The mixed-resolution 1024-anchor pool is also slightly easier than separate within-resolution galleries for some image-side directions, so Table~\ref{tab:retrieval_base} should not be read as a pure single-scale difficulty estimate; equal gallery size does not imply equal difficulty across resolutions. We therefore treat the per-resolution results as a robustness check rather than a ranking of scales. Per-city, spatially blocked, and leave-city-out evaluation, multiple seeds, parent/child visual context, a \texttt{CLS}-versus-central-pooling ablation, and full-scale external-baseline adaptation remain future work.

\begin{acks}
This work was supported by the National Science Centre, Poland,
under grant no. 2024/55/D/ST6/03082. 
Generative AI tools were used to assist with language editing, manuscript organization, and the drafting or refactoring of selected code snippets. All scientific claims, experimental results, implementation details, source code, and citations were reviewed and verified by the authors.
\end{acks}

\bibliographystyle{ACM-Reference-Format}
\bibliography{bibliography}



\appendix
\twocolumn[%
\begin{minipage}{\textwidth}
\centering
\captionof{table}{Positioning against related geospatial representation learning approaches. GeoLEON-CLIP is emphasized as the reference model.}
\label{tab:related_work_positioning}
\scriptsize
\setlength{\tabcolsep}{3pt}
\renewcommand{\arraystretch}{1.00}
\begin{tabularx}{\textwidth}{@{}p{0.15\textwidth}p{0.22\textwidth}p{0.21\textwidth}X@{}}
\toprule
\textbf{Work} & \textbf{Modalities} & \textbf{Training signal} & \textbf{Relation to the reference model} \\
\midrule
Hex2Vec \cite{wozniak2021hex2vec}
& OSM tags, H3 cells
& Spatial-context prediction
& Learns OSM-based cell embeddings, but does not align imagery or text. \\
S2Vec \cite{s2vec2026}
& Built-environment features, S2 cells
& Masked autoencoding
& Learns self-supervised spatial embeddings, but remains unimodal. \\
SatCLIP / GeoCLIP \cite{klemmer2025satclip,vivancocepeda2023geoclip}
& Imagery, location
& Image--location contrastive learning
& Learns location-aware embeddings without an explicit OSM graph branch. \\
RemoteCLIP / GeoRSCLIP \cite{liu2024remoteclip,zhang2024rs5m}
& Remote-sensing imagery, text
& Vision--language contrastive learning
& Aligns images and text, but does not model map structure as a graph modality. \\
SkyScript \cite{wang2024skyscript}
& Remote-sensing imagery, OSM-derived text
& Vision--language pretraining
& Uses OSM to generate textual semantics; GeoLEON-CLIP keeps OSM as a graph modality. \\
UrbanCLIP \cite{yan2024urbanclip}
& Urban imagery, generated text
& Image--text contrastive learning
& Closest urban CLIP-style related work, but lacks graph-anchored tri-modal alignment. \\
\midrule
\textbf{GeoLEON-CLIP (ours)}
& \textbf{OSM graph, satellite imagery, text}
& \textbf{Tri-modal contrastive learning}
& \textbf{Reference model: an OSM/H3 graph branch anchors visual and textual region semantics.} \\
\bottomrule
\end{tabularx}
\end{minipage}
\vspace{0.6em}
]

\section{Positioning Against Related Work}
\label{app:positioning}

Table~\ref{tab:related_work_positioning} places GeoLEON-CLIP among the closest
spatial and multimodal representation learners. Hex2Vec and S2Vec learn useful
map-derived region embeddings but remain unimodal. SatCLIP and GeoCLIP align
imagery with location, while RemoteCLIP, GeoRSCLIP, SkyScript, and UrbanCLIP
connect imagery with language. GeoLEON-CLIP differs by retaining OSM as an
explicit graph modality and using it to organize both visual and textual region
semantics in the same H3-indexed space.

\section{OpenStreetMap Categories}
\label{app:osm-categories}

Table~\ref{tab:osm_categories} lists the 31 tags aggregated into graph node
features. Points are assigned by location; ways and relations use their
centroids.
\begin{minipage}[t]{\columnwidth}

\centering
\captionof{table}{OpenStreetMap categories used for graph node features.}
\label{tab:osm_categories}
\footnotesize
\setlength{\tabcolsep}{3pt}
\renewcommand{\arraystretch}{1.10}
\begin{tabularx}{\columnwidth}{@{}p{0.29\columnwidth}X@{}}
\toprule
\textbf{Domain} & \textbf{OSM tags} \\
\midrule
Health \& Wellness & \texttt{amenity=hospital}, \texttt{amenity=pharmacy} \\
Services & \texttt{amenity=bank}, \texttt{amenity=police} \\
Shops & \texttt{shop=supermarket}, \texttt{shop=bakery}, \texttt{shop=greengrocer}, \texttt{shop=alcohol}, \texttt{shop=clothes} \\
Food \& Drink & \texttt{amenity=restaurant}, \texttt{amenity=bar}, \texttt{amenity=nightclub} \\
Tourism & \texttt{tourism=hotel}, \texttt{tourism=museum} \\
\midrule
Land Use & \texttt{landuse=cemetery}, \texttt{landuse=industrial} \\
Leisure & \texttt{leisure=park}, \texttt{leisure=sports\_centre}, \texttt{leisure=playground} \\
Buildings & \texttt{building=office}, \texttt{building=house}, \texttt{building=apartments} \\
Transport & \texttt{railway=station}, \texttt{railway=tram\_stop}, \texttt{highway=bus\_stop}, \texttt{aeroway=aerodrome} \\
Culture and Education & \texttt{amenity=cinema}, \texttt{amenity=theatre}, \texttt{amenity=library}, \texttt{amenity=place\_of\_worship}, \texttt{amenity=school} \\
\bottomrule
\end{tabularx}
\end{minipage}
\par\medskip

\section{Chained Recall Budget Grids}
\label{app:chained-grids}

Table~\ref{tab:chained_grids} extends Table~\ref{tab:chained_recall} to all
tested first- and second-hop budgets. Recall increases monotonically along both
axes, but equal products $AB$ do not imply equivalent candidate sets because
the graph neighbors selected at the first hop determine what can be reached at
the second. For Image--Graph--Text, every above-diagonal entry exceeds its
transpose, showing that a wider second hop is consistently more productive
than retrieving more graph cells first. The reverse chain is less asymmetric:
a wider second hop is preferable at small budgets, whereas a broader first hop
often becomes more useful at moderate budgets. 

\begin{minipage}[t]{\columnwidth}

The differences are nevertheless
small for several mirrored allocations. At $AB=4$, the balanced $2@2$ setting
outperforms both $1@4$ and $4@1$ in both directions. The largest tested setting,
$10@10$, reaches 94.4\% and 95.6\% recall. Relative to matched-budget direct
image--text retrieval, the chained route has a clear advantage only at $1@1$,
where it improves recall by 4.8 and 4.4 percentage points.

\centering
\captionof{table}{Chained recall $\mathrm{R}@A@B$ (\%). Rows give first-hop budget $A$ and columns second-hop budget $B$.}
\label{tab:chained_grids}
\footnotesize
\setlength{\tabcolsep}{3.5pt}
\renewcommand{\arraystretch}{1.12}
\begin{tabular*}{\columnwidth}{@{\extracolsep{\fill}}l r r r r r r@{}}
\toprule
\multicolumn{7}{c}{\itshape Image--Graph--Text} \\
\cmidrule(lr){1-7}
$A\backslash B$ & 1 & 2 & 3 & 4 & 5 & 10 \\
\midrule
1  & \textbf{66.5} & 72.6 & 74.7 & 76.9 & 78.6 & 84.2 \\
2  & 71.2 & 77.6 & 79.9 & 82.2 & 84.1 & 88.9 \\
3  & 72.8 & 79.3 & 81.5 & 83.7 & 85.4 & 90.3 \\
4  & 73.9 & 80.7 & 82.8 & 85.0 & 86.5 & 91.6 \\
5  & 74.2 & 81.4 & 83.6 & 85.9 & 87.6 & 92.7 \\
10 & 74.9 & 83.1 & 85.4 & 87.9 & 89.6 & 94.4 \\
\bottomrule
\end{tabular*}

\vspace{0.8em}

\begin{tabular*}{\columnwidth}{@{\extracolsep{\fill}}l r r r r r r@{}}
\toprule
\multicolumn{7}{c}{\itshape Text--Graph--Image} \\
\cmidrule(lr){1-7}
$A\backslash B$ & 1 & 2 & 3 & 4 & 5 & 10 \\
\midrule
1  & \textbf{66.0} & 72.7 & 74.0 & 75.1 & 75.8 & 78.3 \\
2  & 70.2 & 77.1 & 79.2 & 80.5 & 81.8 & 85.3 \\
3  & 71.9 & 79.0 & 81.1 & 82.6 & 84.0 & 87.7 \\
4  & 73.6 & 81.2 & 83.5 & 85.5 & 86.7 & 90.3 \\
5  & 74.4 & 82.1 & 84.8 & 86.9 & 87.9 & 92.0 \\
10 & 77.4 & 86.0 & 88.8 & 90.4 & 91.7 & 95.6 \\
\bottomrule
\end{tabular*}
\end{minipage}
\par\medskip

\section{Qualitative Cross-Modal Embedding Arithmetic}
\label{app:cross-modal-arithmetic}

Figure~\ref{fig:cross-modal-arithmetic-all-modalities} shows three qualitative
examples in the shared embedding space. Each combines two projected embeddings
from different modalities and H3 cells, then retrieves a nearest neighbor in
the third modality. Thin frames mark operands, the thicker frame marks the
retrieved modality, and unframed views provide tri-modal context. The examples
illustrate possible compositional structure but do not constitute a quantified
analogy benchmark.

\clearpage


\newlength{\HexBlockW}
\newlength{\ModalityW}
\newlength{\ModalityH}

\setlength{\HexBlockW}{0.282\textwidth}
\setlength{\ModalityW}{0.247\textwidth}
\setlength{\ModalityH}{1.25cm}

\newcommand{\ActiveFrame}[1]{%
  \begingroup
  \setlength{\fboxsep}{1.6pt}%
  \setlength{\fboxrule}{0.45pt}%
  \fbox{#1}%
  \endgroup
}

\newcommand{\ResultFrame}[1]{%
  \begingroup
  \setlength{\fboxsep}{1.8pt}%
  \setlength{\fboxrule}{0.95pt}%
  \fbox{#1}%
  \endgroup
}

\newcommand{\InactiveFrame}[1]{%
  \begingroup
  \setlength{\fboxsep}{1.8pt}%
  \setlength{\fboxrule}{0pt}%
  \fbox{#1}%
  \endgroup
}

\newcommand{\SmallImageCard}[1]{%
  \begin{minipage}[c][\ModalityH][c]{\ModalityW}
    \centering
    {\scriptsize\bfseries image}\\[0.10em]
    \includegraphics[width=0.78\linewidth,height=0.84cm,keepaspectratio]{#1}
  \end{minipage}%
}

\newcommand{\SmallGraphCard}[1]{%
  \begin{minipage}[c][\ModalityH][c]{\ModalityW}
    \centering
    {\scriptsize\bfseries graph}\\[0.10em]
    \includegraphics[width=0.82\linewidth,height=0.84cm,keepaspectratio]{#1}
  \end{minipage}%
}

\newcommand{\SmallTextCard}[1]{%
  \begin{minipage}[c][\ModalityH][c]{\ModalityW}
    \centering
    {\scriptsize\bfseries text}\\[0.18em]
    \begin{minipage}{0.86\linewidth}
      \raggedright\tiny\itshape ``#1''
    \end{minipage}
  \end{minipage}%
}

\newcommand{\HexAllModalities}[7]{%
  \begin{minipage}[c]{\HexBlockW}
    \centering
    {\tiny\ttfamily h3:#1}\\[0.18em]
    #5{\SmallImageCard{#2}}\\[-0.10em]
    #6{\SmallGraphCard{#3}}\\[-0.10em]
    #7{\SmallTextCard{#4}}
  \end{minipage}%
}

\newcommand{\ArithmeticCase}[5]{%
  \begin{minipage}{0.99\textwidth}
    \centering
    {\small\bfseries #1}\\[-0.15em]
    {\scriptsize\ttfamily #2}\\[0.25em]
    {\renewcommand{\arraystretch}{0.96}%
    \begin{tabular}{@{}c@{\hspace{0.25em}}c@{\hspace{0.25em}}c@{\hspace{0.25em}}c@{\hspace{0.25em}}c@{}}
      #3 & {\Large $+$} & #4 & {\Large $\Longrightarrow$} & #5
    \end{tabular}}
  \end{minipage}
  \par\vspace{0.35em}
}


\begin{figure*}[p]
  \centering
  \begin{adjustbox}{max totalsize={\textwidth}{0.90\textheight},center}
  \begin{minipage}{\textwidth}
  \centering

  \ArithmeticCase
  {Case 1: graph + text retrieves image}
  {graph Beijing + text Sao Paulo $\Rightarrow$ image Sao Paulo}
  {%
    \HexAllModalities
    {873181809ffffff}
    {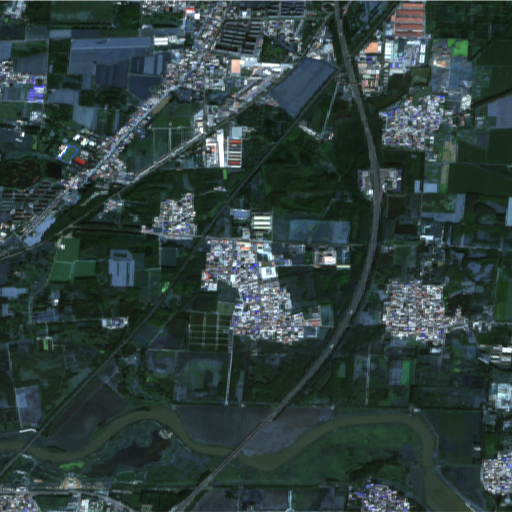}
    {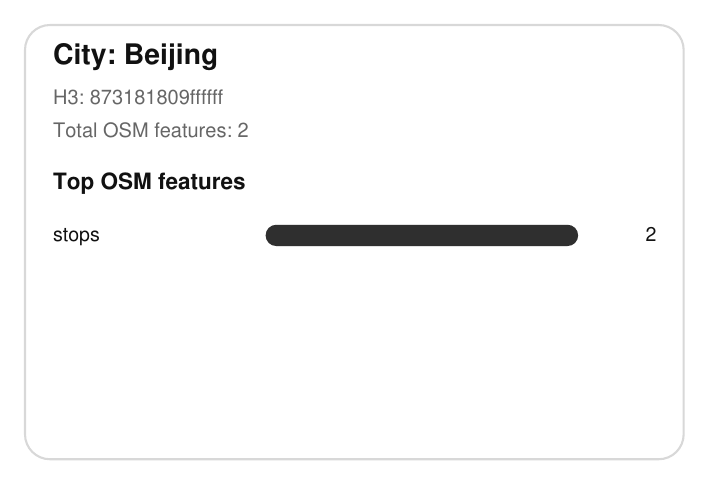}
    {A sparse, undeveloped, or purely residential area with minimal documented infrastructure or commercial activity.}
    {\InactiveFrame}
    {\ActiveFrame}
    {\InactiveFrame}
  }
  {%
    \HexAllModalities
    {87a810000ffffff}
    {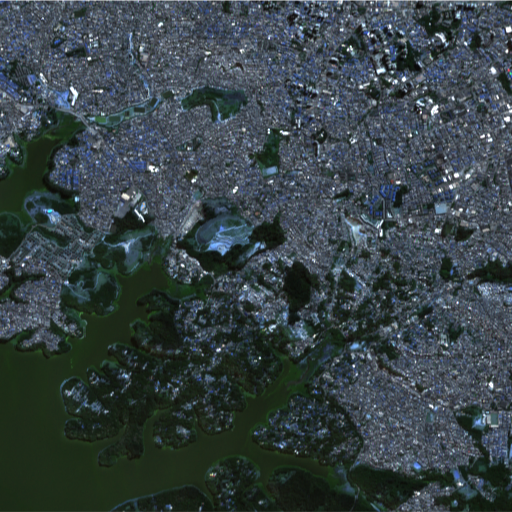}
    {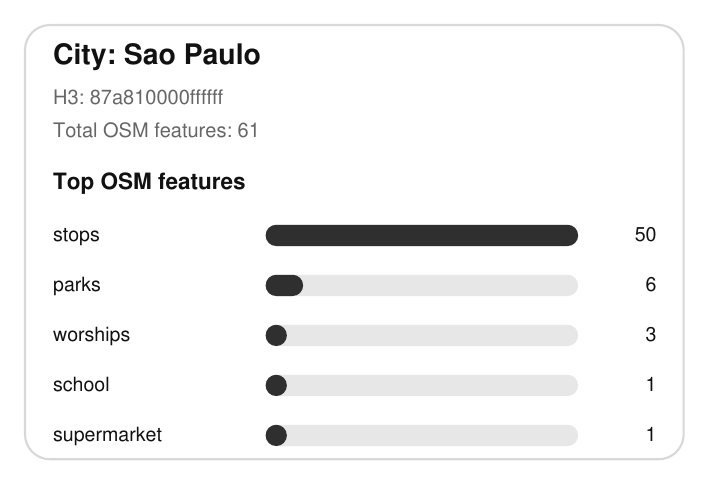}
    {The urban district in question boasts a vibrant atmosphere characterized by its diverse array of religious sites and cultural gatherings.}
    {\InactiveFrame}
    {\InactiveFrame}
    {\ActiveFrame}
  }
  {%
    \HexAllModalities
    {87a810146ffffff}
    {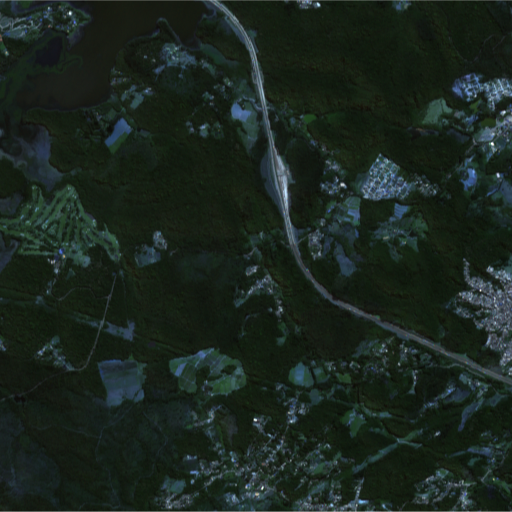}
    {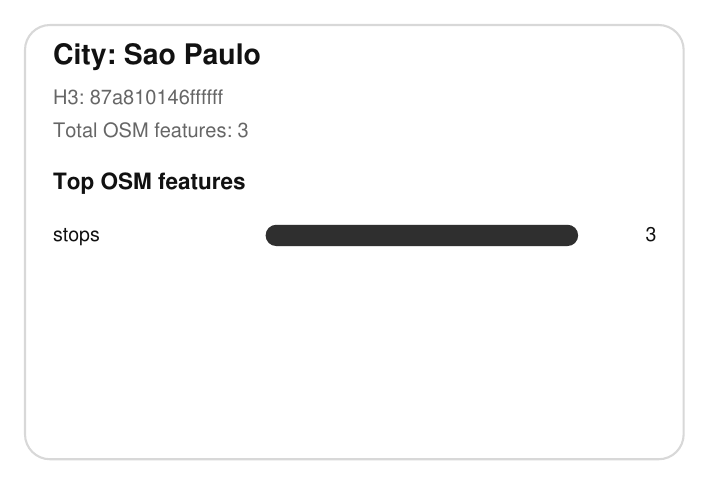}
    {A sparse, undeveloped, or purely residential area with minimal documented infrastructure or commercial activity.}
    {\ResultFrame}
    {\InactiveFrame}
    {\InactiveFrame}
  }

  \ArithmeticCase
  {Case 2: graph + image retrieves text}
  {graph Moscow + image Tokyo $\Rightarrow$ text Tokyo}
  {%
    \HexAllModalities
    {871181266ffffff}
    {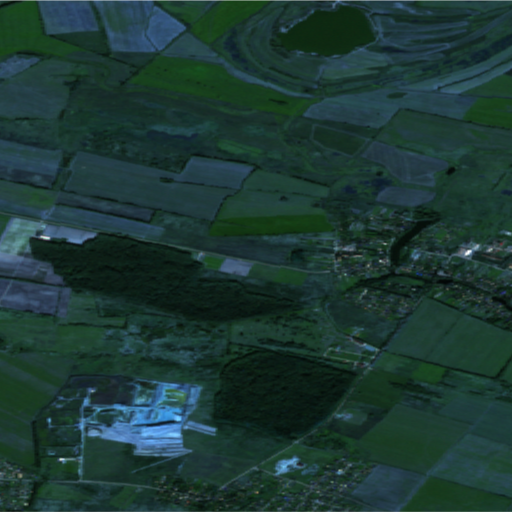}
    {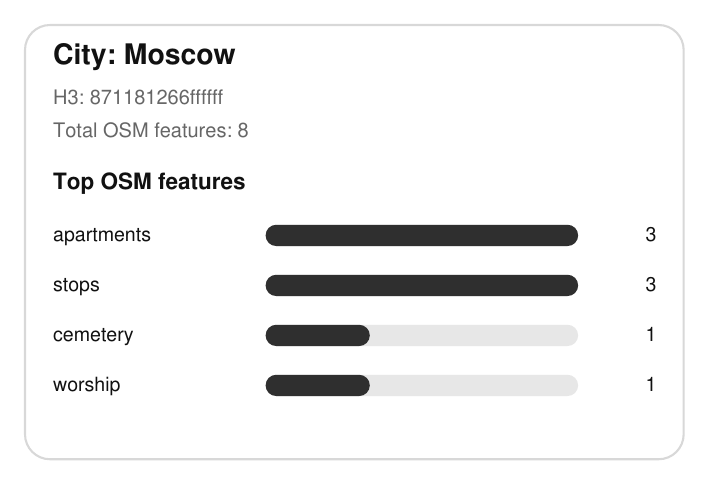}
    {The urban district is characterized by its diverse array of religious sites and residential properties, with a notable concentration of public transportation stops.}
    {\InactiveFrame}
    {\ActiveFrame}
    {\InactiveFrame}
  }
  {%
    \HexAllModalities
    {872f5840bffffff}
    {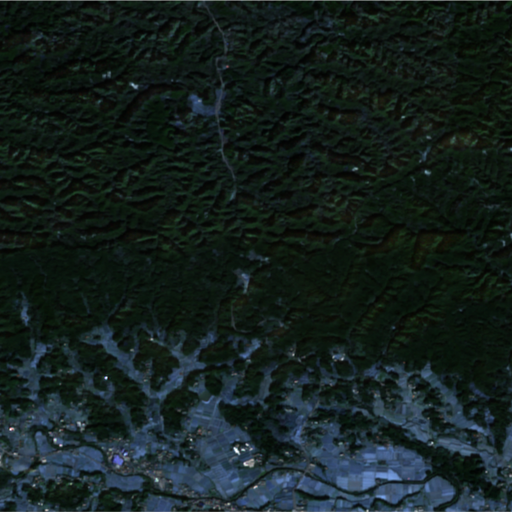}
    {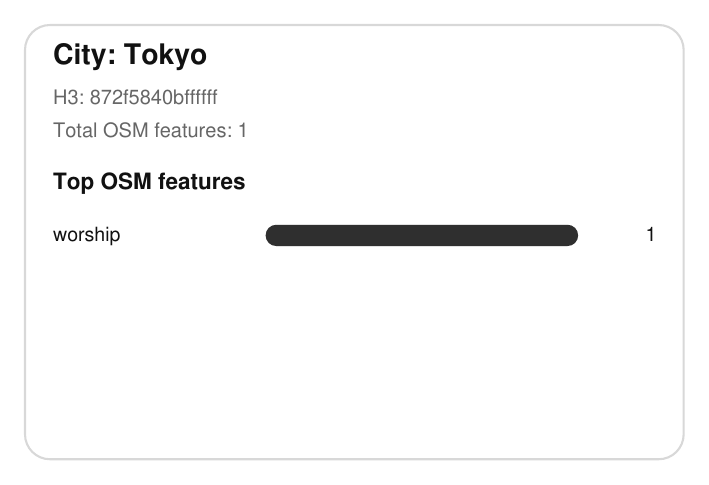}
    {A sparse, undeveloped, or purely residential area with minimal documented infrastructure or commercial activity.}
    {\ActiveFrame}
    {\InactiveFrame}
    {\InactiveFrame}
  }
  {%
    \HexAllModalities
    {872f5a48cffffff}
    {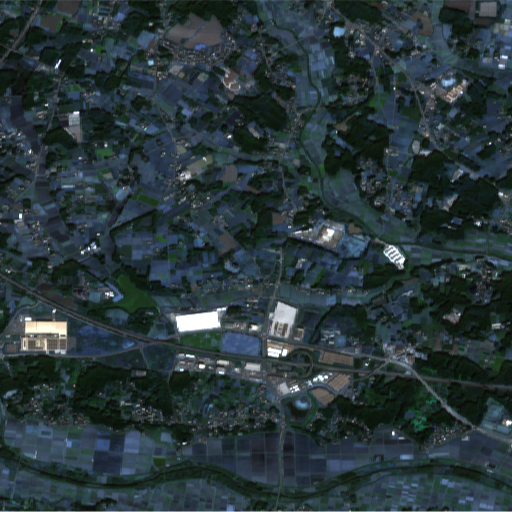}
    {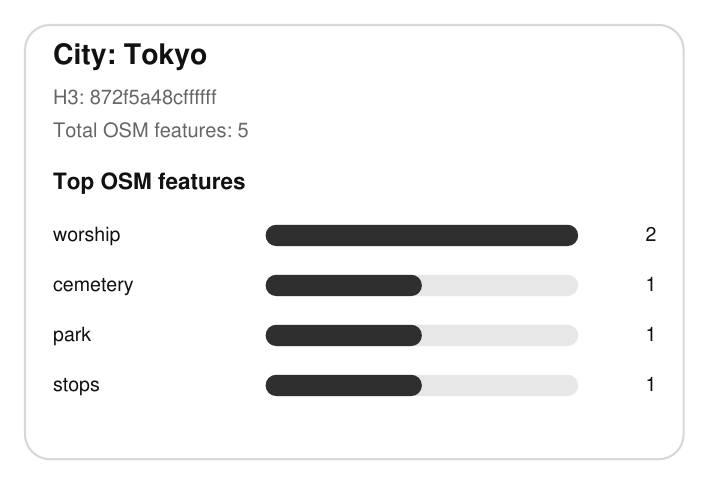}
    {The urban district is characterized by its blend of religious sites, public transportation hubs, and natural green spaces, creating a unique atmosphere.}
    {\InactiveFrame}
    {\InactiveFrame}
    {\ResultFrame}
  }

  \ArithmeticCase
  {Case 3: image + text retrieves graph}
  {image Berlin + text Shanghai $\Rightarrow$ graph Shanghai}
  {%
    \HexAllModalities
    {871f18840ffffff}
    {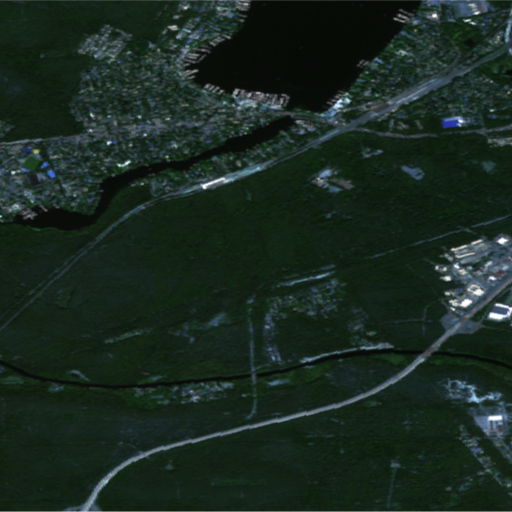}
    {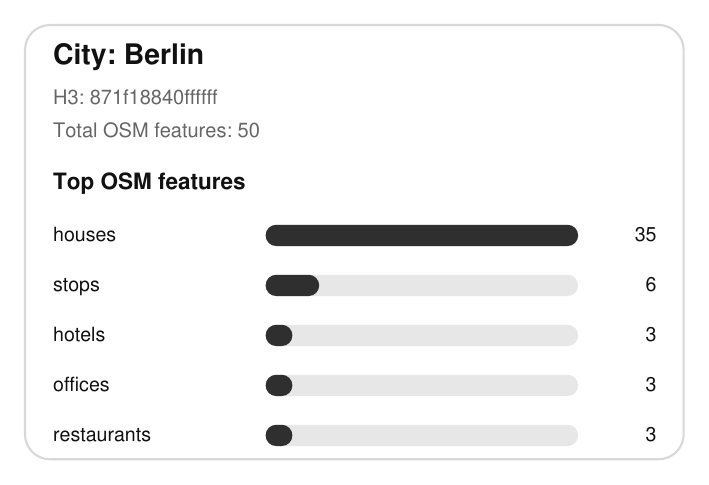}
    {The urban district boasts a vibrant mix of amenities that reflect its diverse community. Nestled between two major avenues, the area is home to several establishments catering to various tastes.}
    {\ActiveFrame}
    {\InactiveFrame}
    {\InactiveFrame}
  }
  {%
    \HexAllModalities
    {873098000ffffff}
    {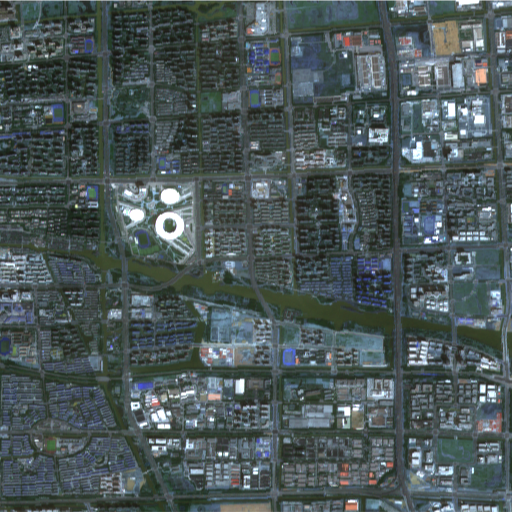}
    {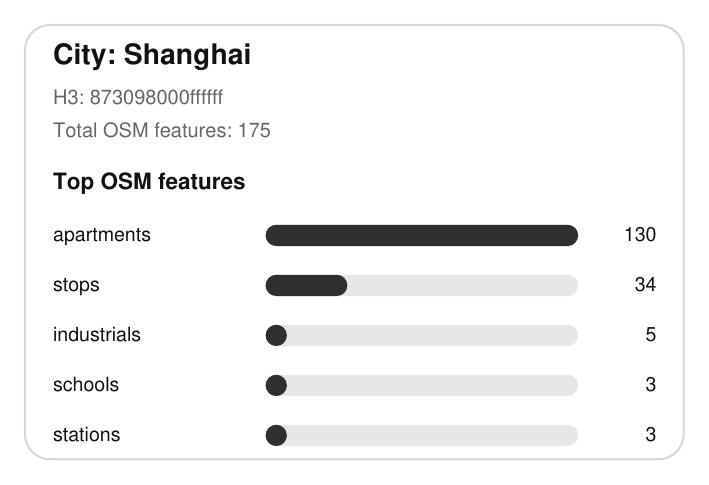}
    {The urban district boasts a vibrant mix of amenities that reflect its dynamic character, with hospitals, schools, and local services.}
    {\InactiveFrame}
    {\InactiveFrame}
    {\ActiveFrame}
  }
  {%
    \HexAllModalities
    {87309d0a2ffffff}
    {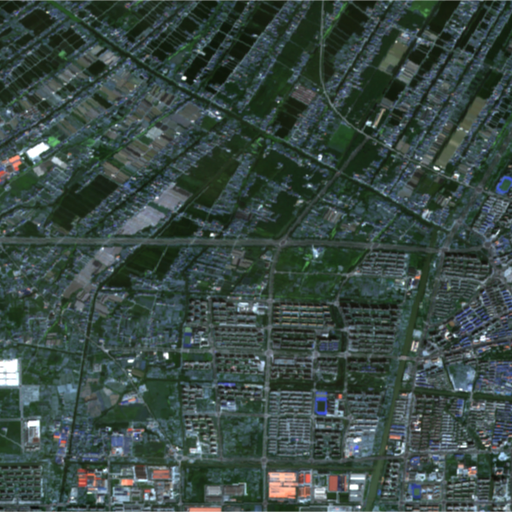}
    {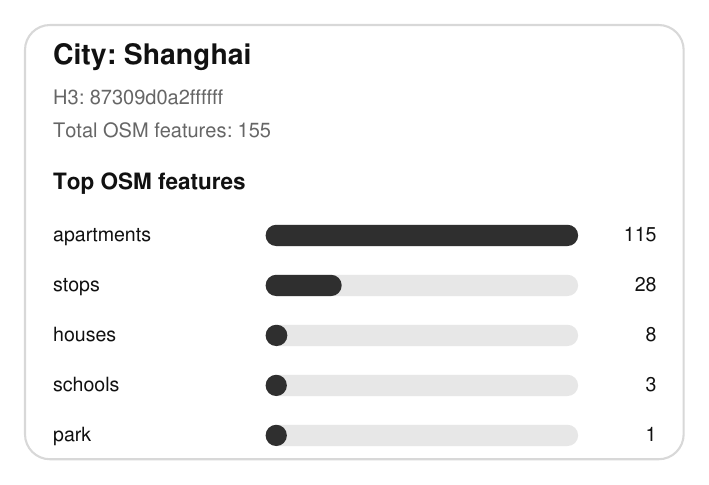}
    {The urban district is characterized by its vibrant mix of residential and educational spaces, with high population density.}
    {\InactiveFrame}
    {\ResultFrame}
    {\InactiveFrame}
  }

  \end{minipage}
  \end{adjustbox}
  \caption{Qualitative examples of cross-modal embedding arithmetic in the
  aligned GeoLEON-CLIP space. For each H3 cell, all three modalities are shown:
  satellite image, graph-derived region representation, and text description.
  Thin frames indicate the two modalities used as operands in the arithmetic
  expression. The thicker frame indicates the retrieved target modality.}
  \Description{Three qualitative cross-modal arithmetic examples. Each row shows two H3 cells combined with a plus sign and a retrieved H3 cell after an implication arrow. Every H3 cell displays an image, a graph card, and a text description; operands are outlined with thin frames and retrieval outputs with thicker frames.}
  \label{fig:cross-modal-arithmetic-all-modalities}
\end{figure*}

\end{document}